\documentclass{article}

\usepackage{algorithm}  
\usepackage{algorithmicx}  
\usepackage{algpseudocode}

\usepackage{hyperref}
\usepackage{url}
\usepackage[utf8]{inputenc} 
\usepackage[T1]{fontenc}    
\usepackage{booktabs}       
\usepackage{nicefrac}       
\usepackage{microtype}      
\usepackage{amsfonts,amsmath,amssymb,amsthm}
\usepackage{dsfont}
\usepackage{lipsum}
\usepackage{wrapfig}
\usepackage{subfig}
\usepackage{enumitem}
\usepackage{enumerate}
\usepackage{xspace}

\usepackage[flushleft]{threeparttable}
\usepackage{vcell}
\usepackage{makecell}
\usepackage{multirow}
\usepackage{multicol}
\usepackage{xcolor}
\usepackage{tikz}
\usepackage[font=small]{caption}
\usepackage{booktabs}
\usepackage{colortbl}

\usepackage{listings}

\usepackage{tcolorbox}
\definecolor{mine}{RGB}{205, 232, 248}%
\definecolor{my_g}{RGB}{128, 128, 128}

\usepackage{amssymb}
\usepackage{amsmath}
\usepackage{comment}
\usepackage{amsthm}

\theoremstyle{plain}
\newtheorem{theorem}{Theorem}[section]

\newtheorem{definition}[theorem]{Definition}

\definecolor{dkgreen}{rgb}{0,0.6,0}
\definecolor{gray}{rgb}{0.5,0.5,0.5}
\definecolor{mauve}{rgb}{0.58,0,0.82}

\usepackage{hyperref}
\usepackage{cleveref}
\Crefname{equation}{Eqn.}{Eqns.}
\Crefname{figure}{Fig.}{Figs.}
\Crefname{figure*}{Fig.}{Figs.}
\Crefname{tabular}{Tab.}{Tabs.}

\usepackage{subfig}
\usepackage{graphicx}
\usepackage{amsmath}
\usepackage{bbm}

\usepackage[accepted]{icml2021}

\icmltitlerunning{Offline Reinforcement Learning with Imbalanced Datasets}

\begin{document}

\twocolumn[
\icmltitle{Offline Reinforcement Learning with Imbalanced Datasets}




\begin{icmlauthorlist}
\icmlauthor{Li Jiang}{thu-sigs}
\icmlauthor{Sijie Cheng}{thu-cs,thu-air}
\icmlauthor{Jielin Qiu}{cmu}
\icmlauthor{Haoran Xu}{thu-air}
\icmlauthor{WaiKin Chan}{thu-sigs}
\icmlauthor{Ding Zhao}{cmu}
\end{icmlauthorlist}

\icmlaffiliation{thu-cs}{Deptartment of Computer Science and Technology, Tsinghua University}
\icmlaffiliation{thu-sigs}{Shenzhen International Graduate School, Tsinghua University}
\icmlaffiliation{thu-air}{Institute for AI Industry Research, Tsinghua University}
\icmlaffiliation{cmu}{Carnegie Mellon University}

\icmlcorrespondingauthor{Li Jiang}{jiangli3859@gmail.com}

\icmlkeywords{Machine Learning, ICML, Deep Reinforcement Learning, Imitation Learning, Batch Reinforcement Learning, Off-Policy}

\vskip 0.3in
]



\printAffiliationsAndNotice{}  

\begin{abstract}
The prevalent use of benchmarks in current offline reinforcement learning (RL) research has led to a neglect of the imbalance of real-world dataset distributions in the development of models. The real-world offline RL dataset is often imbalanced over the state space due to the challenge of exploration or safety considerations. In this paper, we specify properties of imbalanced datasets in offline RL, where the state coverage follows a power law distribution characterized by skewed policies. Theoretically and empirically, we show that typically offline RL methods based on distributional constraints, such as conservative Q-learning (CQL), are ineffective in extracting policies under the imbalanced dataset. Inspired by natural intelligence, we propose a novel offline RL method that utilizes the augmentation of CQL with a retrieval process to recall past related experiences, effectively alleviating the challenges posed by imbalanced datasets. We evaluate our method on several tasks in the context of imbalanced datasets with varying levels of imbalance, utilizing the variant of D4RL. Empirical results demonstrate the superiority of our method over other baselines.

\end{abstract}

\section{Introduction}
Offline reinforcement learning (RL), known as batch RL, \citep{lange2012batch, levine2020offline} holds great promise in learning high-quality policies from previously logged datasets, without further interaction with the environment and collect trajectories that may be dangerous or expensive \citep{tang2021model, kalashnikov2021mt, zhan2022deepthermal}. Such promise makes real-world RL more realistic and enables better generalization abilities by incorporating diverse previous experiences \citep{levine2020offline}. 

To mitigate the fundamental challenge offline RL, i.e., distributional shift \citep{levine2020offline}, most current studies generally enforce pessimism on policy updates \citep{fujimoto2019off, wu2019behavior, kumar2019Stabilizing,fujimoto2021minimalist}, or value updates \citep{kumar2020conservative, an2021uncertainty, bai2022pessimistic}. The pessimism explicitly or implicitly constrains the learned policy to the behavior policy and has been shown to be effective in standard benchmarks in \citep{fu2020d4rl, gulcehre2020rl, qin2021neorl}. However, it should be noted that while real-world distributions are rarely uniform, the datasets used in these benchmarks often feature near-uniform state coverage and near-balanced policies. While the imbalanced dataset has been widely studied in supervised learning \citep{buda2018systematic, liu2019large}, it remains under-explored in the offline RL community even though the static dataset is the only source to extract policies. Real-world datasets in offline RL are often imbalanced mainly due to the challenge of exploration or safety consideration, consisting of skewed policies.

Real-world datasets in offline RL exhibit imbalanced and follow Zipf's law \citep{gabaix1999zipf, newman2005power} across the entire state space, with skewed policies (see \Cref{sec3: current offline RL methods fail}). The states in the dataset from sufficient to insufficient coverage are characterized by \textit{varying policies from mixture to expert ones}. As an example, for safety considerations in industrial control systems, certain adjustments to key parameters may be rare but near-optimal, while other trivial adjustments may be more common but inferior.


Our study shows that most of the existing offline RL algorithms perform poorly when faced with imbalanced datasets. We then provide a provable explanation for this failure, which unfolds in two dimensions. Firstly, \textit{state-agnostic pessimism} in most current approaches fail to guarantee policy improvement. Secondly, \textit{uniform transition sampling} during the course of training may lead to large temporal-difference (TD) errors in off-policy evaluation for states with poor coverage due to inefficient sampling. Ironically, our empirical investigation exposes that re-sampling methods like Prioritized Experience Replay (PER, as outlined in \cite{schaul2015prioritized}) may lead to worse performance in imbalanced offline RL datasets, serving as an amplifier for the distributional shift problem.  

Motivated by natural intelligence, which avoids forgetting rare phenomena by recalling related information from past experiences \citep{mcclelland1995there, leake1996case}, we introduce a new offline RL algorithm to overcome those limitations by augmenting standard offline RL algorithms, e.g., CQL \citep{kumar2020conservative},  with a \textit{retrieval process}. Our proposed method, retrieval-based CQL (RB-CQL), enables agents to effectively utilize related experiences through nearest-neighbor matching from diverse and large-scale datasets, and to directly inform their actions, particularly for states with poor coverage. Our algorithm is evaluated on a series of tasks with imbalanced datasets in the variant of D4RL \citep{fu2020d4rl} and shows the competition compared with state-of-the-art offline RL methods. Note that RB-CQL represents a straightforward approach to alleviate offline RL in the context of real-world imbalanced datasets. We hope this work will inspire future research on real-world RL systems utilizing practical datasets, and serve as a foundation for large-scale offline RL.


\section{Preliminaries}
RL problem is typically characterized by a Markov Decision Process (MDP) \citep{sutton1998introduction}, which is specified by a tuple $\mathcal{M}=\langle \mathcal{S}, \mathcal{A}, P, r, \rho, \gamma \rangle$ consisting of a state space $\mathcal{S}$, an action space $\mathcal{A}$, a transition probability function $P(s'|s,a)$, a reward function $r(s,a) \in \mathbb{R}$, an initial state distribution $\rho$, and the discount factor $\gamma \in [0, 1)$. The object of RL is to extract a policy $\pi: \mathcal{S} \rightarrow \mathcal{A}$ that maximizes the expectation of the sum of discounted reward, known as the return $J(\pi)= \mathbb{E}_{\pi}\left[\sum_{t=0}^{\infty} \gamma^{t} r(s_t, a_t)\right]$. 

Based on approximate dynamic programming, off-policy RL methods typically learn a state-action value function (Q-function), characterized as $Q^{\pi}(s, a): \mathcal{S} \times \mathcal{A} \rightarrow \mathbb{R}$ with policy $\pi$, the discounted return when the trajectory starts with $(s, a)$ and all remaining actions are taken via $\pi$. For a given policy $\pi$, the Q-function can be obtained through the Bellman operator $\mathcal{T}^{\pi}: \mathbb{R}^{\mathcal{S} \times \mathcal{A}} \rightarrow \mathbb{R}^{\mathcal{S} \times \mathcal{A}}$, defined as :
\begin{equation*}
    \left(\mathcal{T}^\pi Q\right)(s, a):=r(s, a)+\gamma \underset{\substack{s^{\prime} \sim P(\cdot \mid s, a) \\ a' \sim \pi(\cdot \mid s')}}{\mathbb{E}}\left[Q\left(s^{\prime}, a'\right)\right],
\end{equation*}
where $Q^{\pi}(s, a)$ is a unique fixed point with the contracted Bellman operator for $\gamma \in [0, 1)$ \citep{bertsekas1995neuro}. The optimal Q-function is obtained via $Q^*= \max_{\pi}Q^{\pi}(s,a)$, which aligns with an optimal policy achieved through greedy action choices. Another important concept is the notion of discounted state occupancy, $d_\pi \in \Delta(\mathcal{S})$, defined as $d_\pi(s):=(1-\gamma) \mathbb{E}_{\pi}\left[\sum_{t=0}^{\infty} \gamma^t \mathbbm{1}\left[s_t=s\right]\right]$, which characterizes a distribution over the states visited by the policy.

We narrow our focus to offline RL settings, which aim to learn an optimal policy from a pre-collected dataset $\mathcal{D}$, generated by some unknown behavior policy $ \beta $. The most critical problem in offline RL is that the training policy $\pi$ may generate out-of-distribution action $a'$ that causes erroneous estimation error of $Q(s', a')$ and propagate through the Bellman update, often leading to catastrophic failure to the policy learning. Most recent offline RL methods alleviate the above issue by introducing pessimism either on policy learning \citep{fujimoto2019off, wu2019behavior, kumar2019Stabilizing} or value learning \citep{kostrikov2021iql, kumar2020conservative, wu2021uncertainty}. Most of those methods can be formulated into a generic formulation of distribution-constrained offline RL algorithms \cite{kumar2020conservative}, either explicitly or implicitly. The goal of it is to maximize the return of the learned behavior policy $\pi$ while minimizing the divergence $D(\pi, \beta)$ weighted by a hyperparameter $\alpha$: 

\begin{equation}
\max _\pi \mathbb{E}_{s \sim d^\pi}\left[J(\pi)-\alpha D\left(\pi, \beta\right)(s)\right].
\label{eqn: policy constrain}
\end{equation}


\textbf{Conservative Q-learning.} \citep{kumar2020conservative} enforces pessimism on the value update, based on the actor-critic framework, which can be written as: 
\begin{align*}
\label{eq:cql_training}
\min_{\theta} \alpha (\mathbb{E}_{s \sim \mathcal{D}, a \sim  \rho} \left[Q_\theta (s, a)\right] - \mathbb{E}_{s \sim \mathcal{D}, a \sim \beta} \left[Q_\theta (s, a)\right]) \\ + \frac{1}{2}\mathbb{E}_{(s, a, s') \sim \mathcal{D}}\left[(Q_{\theta}(s, a) - \mathcal{T}^{\pi}\bar{Q}_{\theta'}(s,a))^2\right].
\end{align*}

The second term is the standard Bellman updates error \citep{mnih2015human, lillicrap2015continuous, fujimoto2019off} and $\bar{Q}_{\theta}$ represents the target Q-value with parameters $\theta'$, which is updated to the current network with parameters $\theta$ by averaging $\theta' \leftarrow \tau \theta + (1-\tau)\theta'$ for some small $\tau$. Instead of giving pessimism on policy learning by choosing in-distribution actions, CQL enforces pessimism on value learning (the first term) by minimizing the Q-values under a specific policy distribution $\rho$, which is characterized by the high Q-values, and balancing by maximizing the Q-values of from the behavior policy $\beta$. Note that CQL implicitly constrains the learned policy $\pi$ over the behavior policy $\beta$ (see Eq. (13) in \cite{kumar2020conservative}).

The hyperparameter $\alpha$ controls the degree of pessimism. A high value of $\alpha$ corresponds to a high level of pessimism, indicating that the optimal policy is highly similar to the behavior policy $\beta$, while a low value of $\alpha$ corresponds to a low level of pessimism. The pessimism level, i.e., $\alpha$, is a constant for every state-action pair, and thus the value function learned by CQL is changed to the same extent at all possible state-action pairs. However, the consistent divergence between the learned policy and the behavior policy over all possible state-action pairs may ultimately lead to a decline in the performance of the learned policy and even result in failure in imbalanced datasets. We illustrate these challenges by providing a motivating example and practical experiences in the next session.

\section{Current Offline RL Methods fail in the Imbalanced Dataset}
\label{sec3: current offline RL methods fail}
An imbalanced dataset over the state space in offline RL can be described as the state visitation of the behavior policy $d_\beta(s)$ in the dataset imbalanced. The state visitation of the behavior policy $d_\beta(s)$ follows a power law distribution \citep{gabaix1999zipf, newman2005power} with exponent $\eta \in [0, \infty]$. The probability distribution for a random variable $X$ is:

\begin{equation}
p(X=x)=\frac{1}{Z} \cdot \frac{1}{x^\eta},
\label{eqn: exponential distribution}
\end{equation}

where $Z$ is a normalizing constant term and $\eta$ determines the degree of imbalance in the imbalanced offline RL dataset. A higher value of $\eta$ exacerbates the imbalance in the state space and renders learning from rare experiences more challenging. Furthermore, rare experiences are of importance as they are likely, or in many cases, derived from expert policies. The significant decline in performance observed in current offline RL algorithms can mainly be attributed to two factors:

\textbf{State-agnostic.} Most current offline RL algorithms are state-agnostic and enforce consistent pessimism for all state-action pairs in the datasets. In the context of an imbalanced dataset, it is crucial for the learned policy to remain similar to the behavior policy, $\beta$, in states with insufficient coverage $d_\beta^{-}(s)$, as these states typically consist of expert demonstrations. Conversely, in states with sufficient coverage, it is necessary for the learned policy to deviate from $\beta$ in order to stitch potential useful trajectories, as these states often consist of a mixture of demonstrations.

\textbf{Uniformly sampling.} In deep Q-learning offline RL methods, transitions are uniformly sampled from the dataset, contributing to Bellman updates errors with respect to the probability of transitions in the dataset:
\begin{equation}
\label{eqn: td_error}
\approx \mathbb{E}_{(s, a, s^{\prime}) \sim \mathcal{D}}\left[\left(Q_\theta(s, a)-\mathcal{T}^\pi \bar{Q}_{\theta^{\prime}}(s, a)\right)^2\right].
\end{equation}

During the course of training, those suboptimal transitions from $d_\beta^{+}(s)$ dominate over those near-optimal transitions from $d_\beta^{-}(s)$. It will cause inefficient sampling to near-optimal but rare transitions from $d_\beta^{-}(s)$, which may lead to inferior estimation over Q-functions, especially for tasks with sparse reward requiring enormous exploration.

In the following sections, we introduce a navigation task and a practical experience on \texttt{Antmaze} \cite{fu2020d4rl} with a varying imbalance to demonstrate how the current state-of-the-art offline RL algorithm, i.e. CQL, fails in imbalanced datasets. We then theoretically reveal the failure of current offline RL algorithms under imbalanced datasets. Experiments details can be found in \Cref{appendix: exp details}.

\subsection{State-Agnostic Pessimism}
\label{subsec: state-agnostic}
\begin{figure*}[tbp]
\centering
\resizebox{1\linewidth}{!}{
		\centering
			\begin{minipage}[b]{0.4\textwidth}
				\includegraphics[width=1\textwidth]{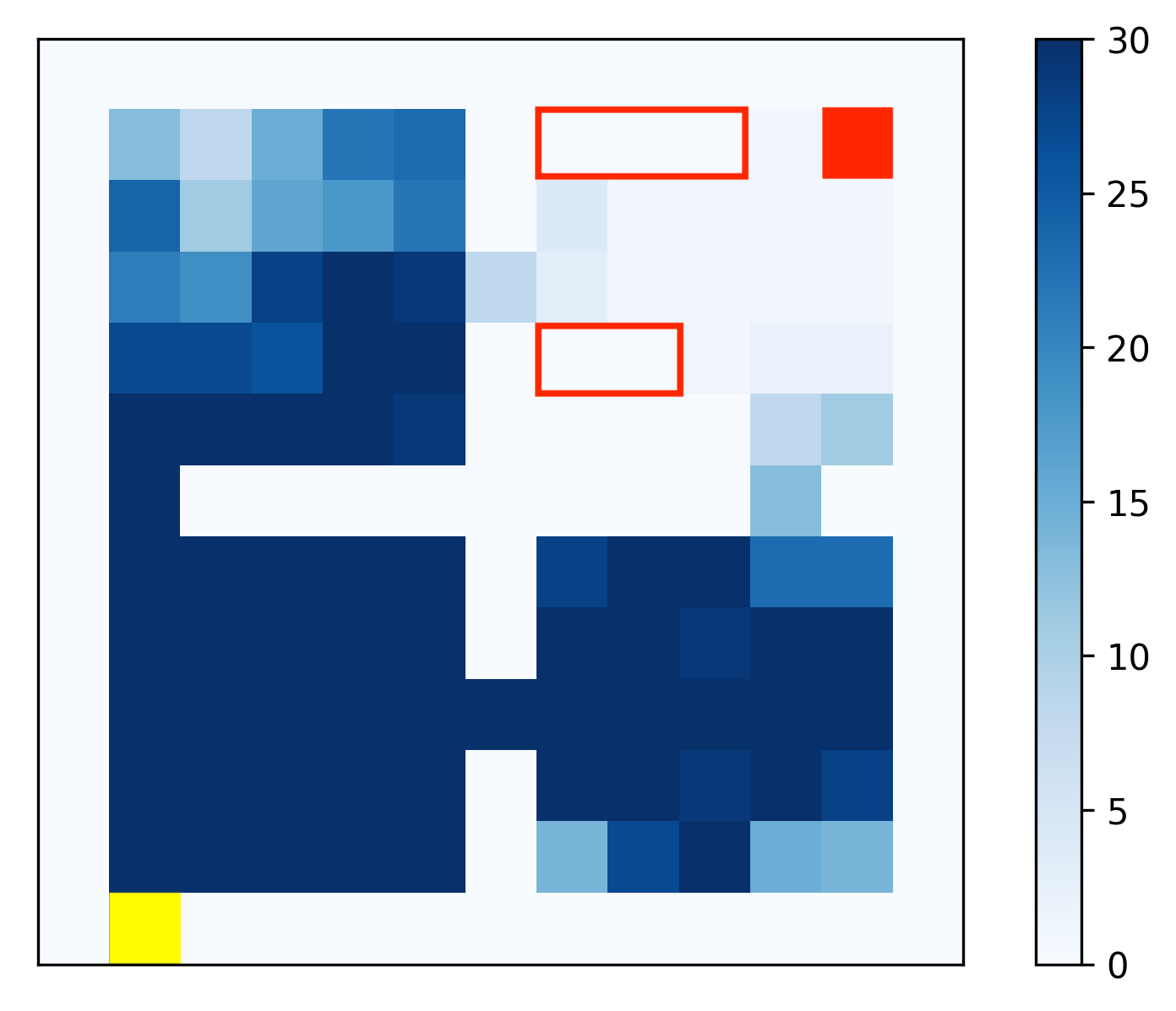}
			\end{minipage}
			\begin{minipage}[b]{0.34\textwidth}
				\includegraphics[width=1\textwidth]{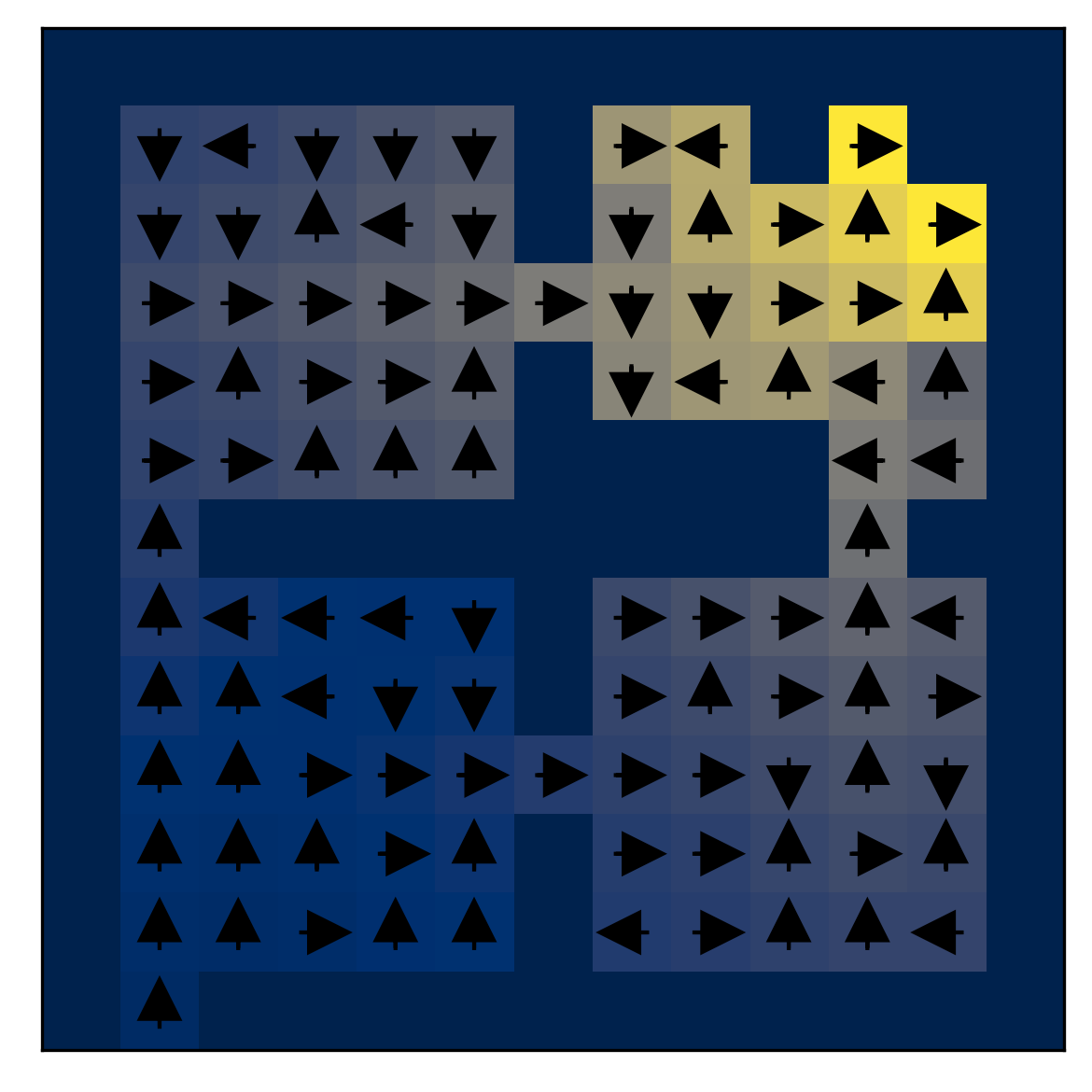}
			\end{minipage}
			\begin{minipage}[b]{0.4\textwidth}
				\includegraphics[width=1\textwidth]{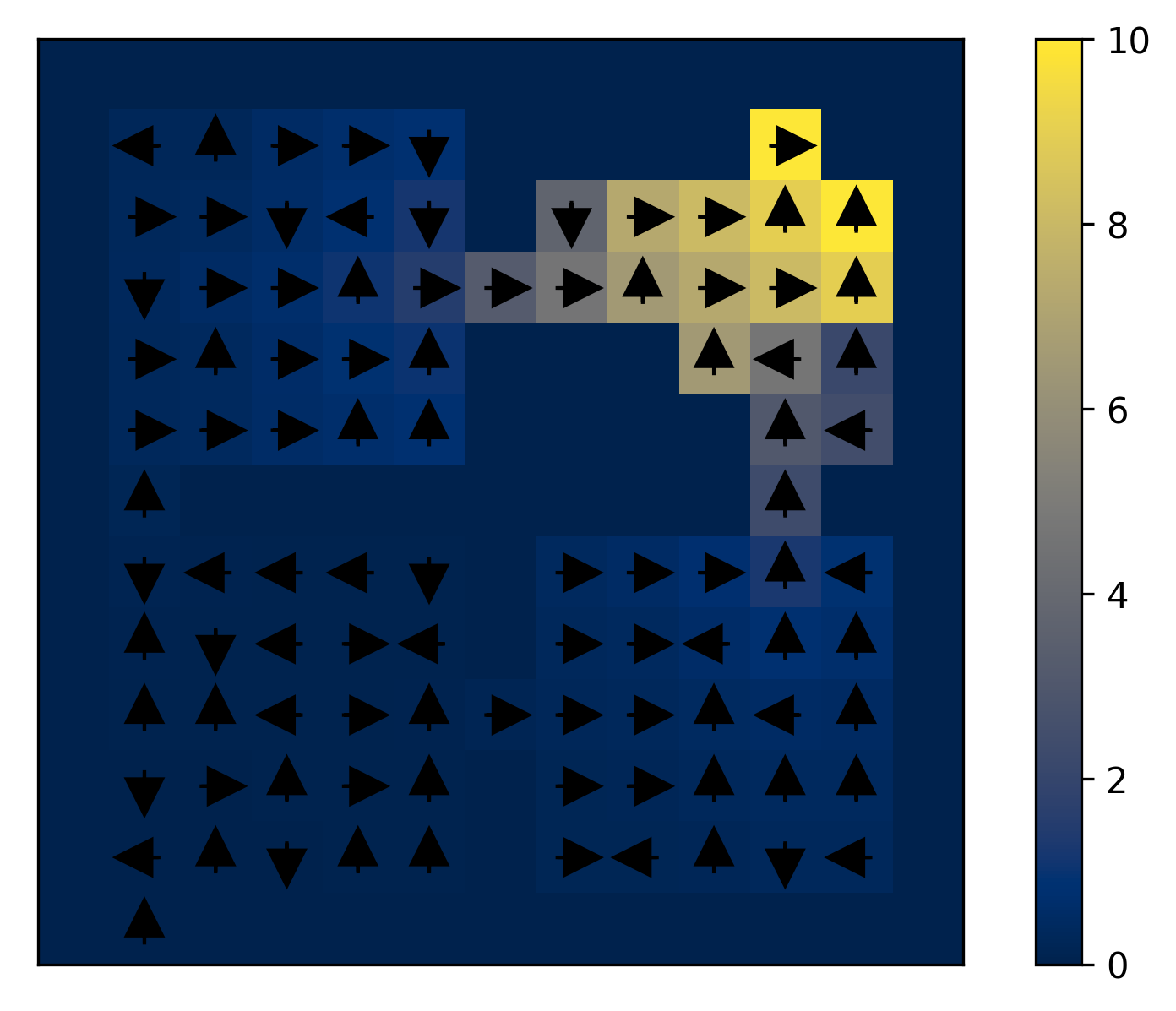}
			\end{minipage}
		}
\caption{\textbf{Left:} The state coverage and task description of the imbalanced dataset in \texttt{Four-room}, where states coverage is featured with the heavy-tail property and states in red rectangle frame have no state coverage. It requires the agent to find a path from the start state (yellow grid) in the first room to the goal state (red grid) in the last room. \textbf{Center:} Policy performance with lower pessimism. The learned policy succeeds in the first room but fails in the last room. 
\textbf{Right:} Policy performance with higher pessimism. The learned policy succeeds in the last room but fails in the first room.}
\label{fig: four-room}
\end{figure*}

We first use a didactic \texttt{Four-room} navigation task in \Cref{fig: four-room} to present the failure of CQL with state-agnostic pessimism. This task exempliﬁes the challenge of exploration with sparse reward and long horizon. The goal of this task is to find a path from the initial state (yellow grid in the first room) to the target state (red grid in the last room) in the grid space, where each room is only connected by one grid with its adjacent rooms. The state and action space are discrete, where the state presents the current location of the agent via the (x, y) coordinate and the action space is 4 corresponding to four orthogonal nearby grids. The agent will receive 10 rewards for reaching the goal state and 0 for all other actions at any feasible state.  

The dataset is generated by a goal-reaching controller from the start to the goal state and the agent executes the correct action with increasing probabilities from 10 $\%$ probability to 100 $\%$. The \texttt{Four-room} dataset is characterized by an imbalanced distribution of state coverage across the entire state space. Specifically, there is a high degree of coverage in the first room, which is accompanied by suboptimal actions, whereas there is insufficient coverage in the last room, where near-optimal policies are present.
The color bar is limited to 30 to better highlight the differences among rare experiences in \Cref{fig: four-room} (left). It should be noted that the samples of some states (e.g., those in the first room) reaching the limit are significantly larger than 30 with a high likelihood.
We illustrate the action choices of two policies trained with CQL, one with low pessimism ($\alpha=5$) and the other with high pessimism ($\alpha=20$), over the entire feasible state space in \Cref{fig: four-room} (center) and (right), respectively.
One feasible state without actions represents that Q-functions are the same in the whole action space, which can be attributed to the missing data  in the original dataset.
The grid color (more saturated to yellow means a higher value) denotes the value function, which is obtained by taking the expectation of all possible actions.

Given a low level of pessimism (i.e., $\alpha=5$), shown in \Cref{fig: four-room} (center), the learned policy is entitled to deviate heavily from the behavior policy, encouraging the agent to stitch suboptimal trajectories and hence successfully get out of the first room. However, low pessimism leads to a poorer estimation of Q-functions in states with insufficient coverage and the failure of the last room. This can be attributed to the high distribution mismatch induced by low pessimism, where insufficient coverage serves as an amplifier to the inaccurate estimation of Q-functions.
On the other hand, in order to succeed in the last room, high pessimism is required to significantly constrain the learned policy to the behavior policy. However, the learned policy fails in the first room (suboptimal trajectories) because of the high similarity to the behavior policy, as shown in \Cref{fig: four-room} (right). 

In order to finish this task, the policy should be imposed state-specific pessimism. In detail, imposing low pessimism for states with sufficient coverage (e.g., states from the first room) and high pessimism for states with insufficient coverage (e.g., states from the last room). Note that lower pessimism is highly risky to out-of-distribution states problem at the test time, resulting in incorrect Q-functions over those states because of the missing in the training dataset.

\subsection{Uniformly Sampling}

To illustrate the failure of current offline RL algorithms, we then train CQL on the imbalanced dataset in AntMaze medium, which is a more sophisticated task with continuous space over states and actions. Similar to \texttt{Four-room} dataset, the expert trajectories reaching the target location in AntmMze datasets are rare and the state coverage is imbalanced. 

\begin{figure}[htbp]
\centering
\resizebox{1\linewidth}{!}{
		\centering
			\begin{minipage}[b]{0.4\textwidth}
				\includegraphics[width=1\textwidth]{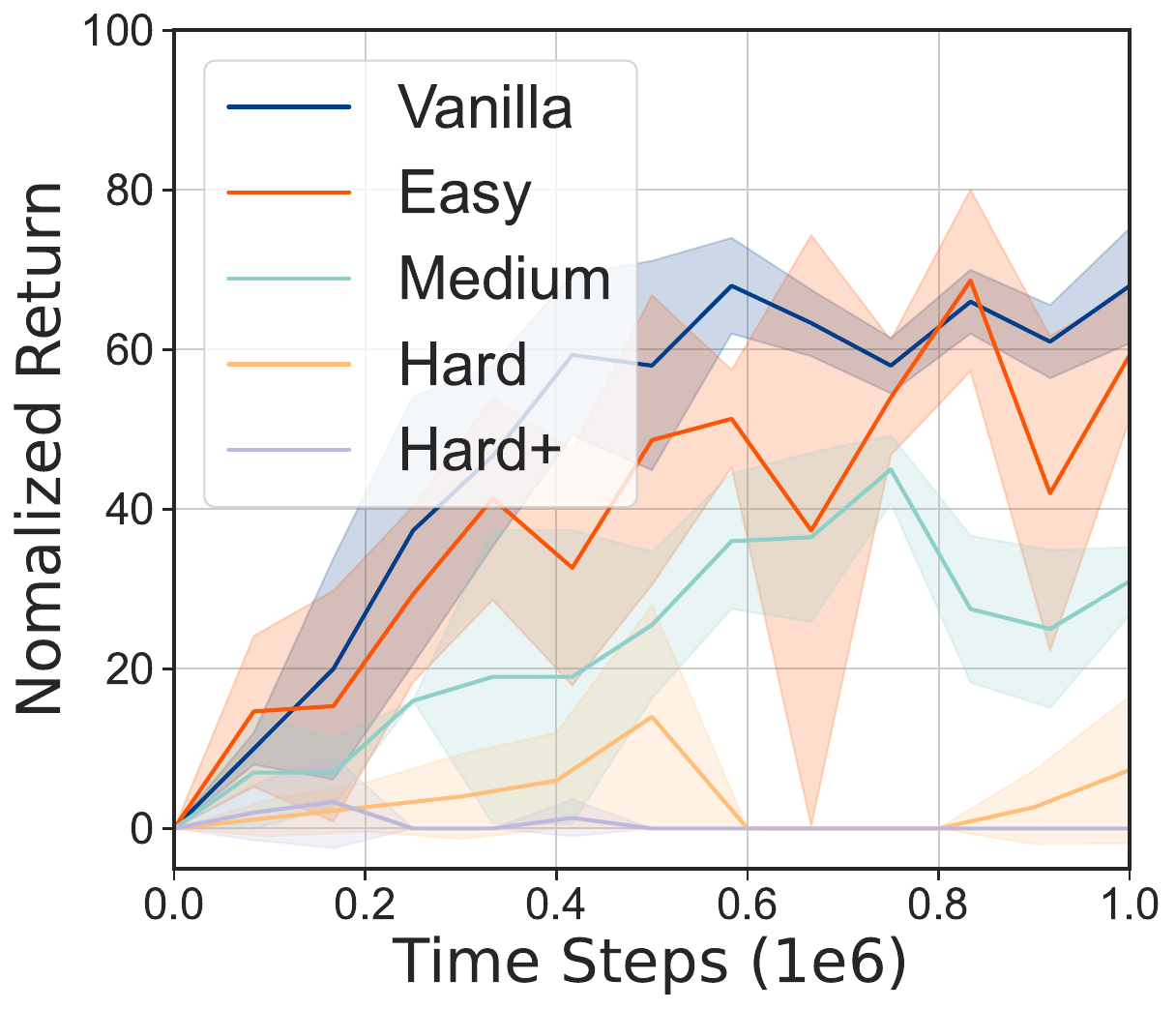}
			\end{minipage}
			\begin{minipage}[b]{0.4\textwidth}
				\includegraphics[width=1\textwidth]{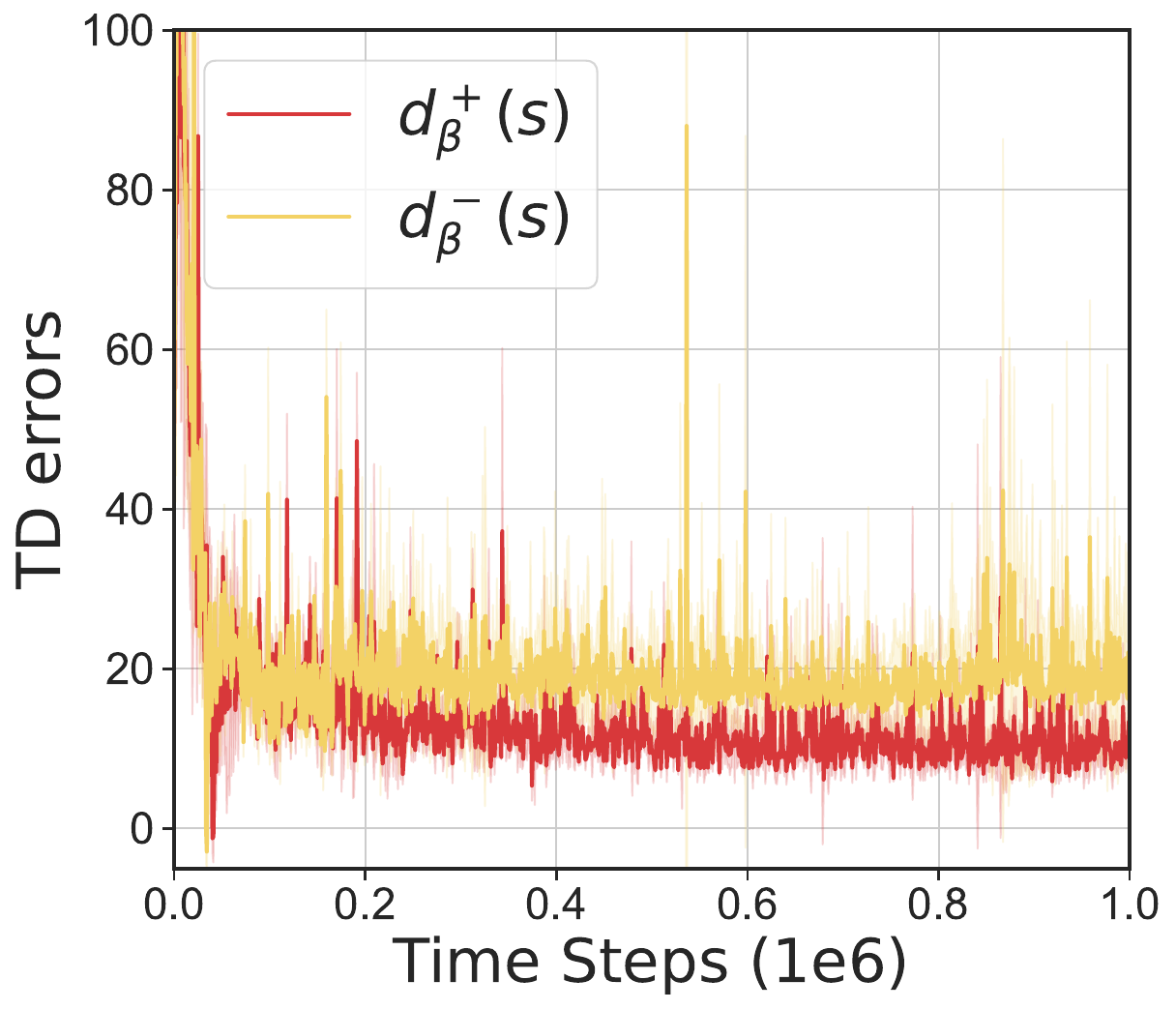}
			\end{minipage}
			\begin{minipage}[b]{0.4\textwidth}
				\includegraphics[width=1\textwidth]{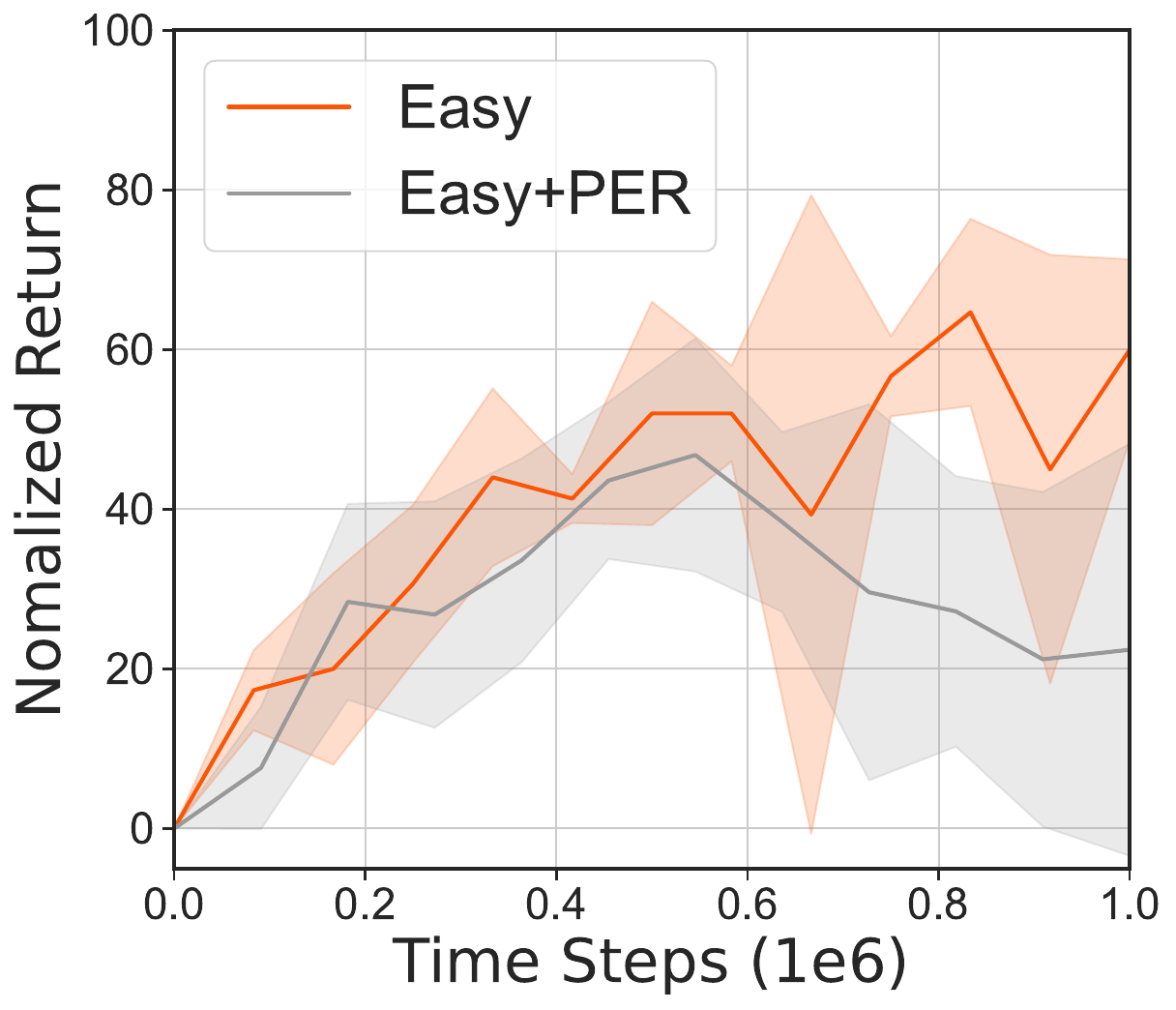}
			\end{minipage}
		}
\caption{\textbf{Left:} The performance of CQL on \texttt{Antmaze-medium} task with imbalanced dataset. With the increasing imbalance of the given dataset from \texttt{Easy} to \texttt{Hard+} (increasing $\alpha$), the performance continues to drop, and even fails in the hard+ dataset. 
\textbf{Center:} TD errors over 500 states from sufficient coverage $d_\beta^+(s)$ and insufficient coverage $d_\beta^-(s)$ on medium-level imbalance, respectively. TD errors from sufficient coverage $d_\beta^+(s)$ are smaller than from insufficient coverage $d_\beta^-(s)$ due to inefficient training.
\textbf{Right:} The performance of CQL and CQL with PER on \texttt{Antmaze-medium} task on easy-level imbalance. PER worsens the final performance as the introduction of additional data distributional shift problem.}
\label{fig: heavy tail with alpha.}
\end{figure}

The imbalance in the dataset is increased with higher $\eta$ and the performance significantly decreases (shown in \Cref{fig: heavy tail with alpha.} (left)). Except for the state-agnostic training process, we hypothesize that uniformly sampling also worsens the performance because of inefficient training on rare expert experiences from $d_\beta^-(s)$, failing to properly estimation of Q-functions. To validate our hypothesis, we track 500 state-action pairs near the target location (i.e., $d_\beta^-(s)$) and the start location (i.e., $d_\beta^+(s)$), respectively. In \Cref{fig: heavy tail with alpha.} (center), we graph the mean of TD errors over and find that TD errors of states from $d_\beta^-(s)$ are larger than states from $d_\beta^+(s)$ around 5. Can we draw techniques from common approaches to the long-tail distribution problem in supervised learning?

Prioritized Experience Replay (PER, as proposed in \citet{schaul2015prioritized}) is a method that modifies the sampling probability of transitions based on their temporal difference (TD) errors. Specifically, it over-samples transitions with high TD errors, potentially adjusting the non-uniformity of the dataset. However, as demonstrated in \Cref{fig: heavy tail with alpha.} (right), re-weighting the loss in \Cref{eqn: td_error} based on TD errors can result in a decrease in overall performance. We posit that the re-sampling process alters the distribution of $d_\beta(s)$, which dynamically shifts the behavior policy $\beta$ with respect to TD errors, and results in an additional divergence between the true $\beta$ and the modified $\beta$.
As a result, the extra distributional shift problem is introduced to the whole training process and yields further incorrect estimations over Q-functions. On the other hand, those techniques are useful in online one-step Q-learning RL methods because the learning policy can collect new experiences and thus avoids the distributional shift problem.

\subsection{Theoretical Analysis}
So far we have empirically seen the continuous performance drop with the increasing imbalance of the imbalanced dataset, now we formally characterize why current pessimism without the consideration of dataset coverage fails to guarantee policy improvement in the imbalanced dataset. In contrast to previous common approaches analysis via concentrability coefficient \citep{liu2020provably, rashidinejad2021bridging}, upper bounding the ratio of state-action visitation a specific policy $d_{\pi}(s, a)$ and the behavior policy from the dataset $d_{\beta}(s)$, i.e., $\max _{s, a} d_\pi(s, a) / d_{\beta}(s) \leq C^\pi$, we use a different metric from the variant of differential concentrability \citep{singh2022offline}, aiming to measure the imbalance of dataset coverage (i.e., the discrepancy of state visitation frequency between adequate and inadequate dataset coverage over the whole state space).

\begin{definition}[Differential concentrability.]
Given a divergence $D$ over the action space, the differential concentrability $C_{\text {diff }}^\pi$ of a given policy $\pi$ with respect to the behavior policy $\beta$ is given by:
\begin{equation}
\underset{\substack{s_1 \sim d_{\beta}(s)^+ \\ s_1 \sim d_{\beta}(s)^-}}{\mathbb{E}}\left[\left(\sqrt{\frac{D\left(\pi, \beta\right)\left(s_1\right)}{d_{\beta}\left(s_1\right)}}-\sqrt{\frac{D\left(\pi, \beta\right)\left(s_2\right)}{d_{\beta}\left(s_2\right)}}\right)^2\right]
\label{eq:  differential concentrability}
\end{equation}
\end{definition}

\Cref{eq: differential concentrability} measures of imbalance in the dataset in the current state-agnostic methods, is characterized by the discrepancy between a given policy $\pi(a|s)$ and the behavior policy $\beta(a|s)$, weighted negatively by the density of states in the denominator, from states with sufficient and insufficient coverage (i.e., $d_{\beta}^+(s)$ and $d_{\beta}^-(s)$), respectively. Instead of considering a simpler scenario where $d_\beta(s) = \text{Unif}(s)$ (i.e., $d_\beta(s_1)=d_\beta(s_2)$) \citep{singh2022offline}, we consider a more realistic scenario where $d_\beta(s_1)>d_\beta(s_2)$, and the imbalance of the dataset is exacerbated by increasing the value of $\eta$ in \Cref{eqn: exponential distribution}. This allows us to examine the effect of imbalanced datasets on the performance of our method.

Under the imbalanced dataset, if the policy divergence $\mathcal{D}$ with respect to any given policy $\pi$ is state-agnostic, whether loose or tight constraints, $C_{\text {diff }}^\pi$ would be large because the denominators are significantly disagreeable (i.e., $d_\beta(s_1)$ is large than $d_\beta(s_2)$). 
However, on the other hand, if we allow $\mathcal{D}$ to be state-specific with respect to the state coverage (i.e., loose constraints to states from $d_{\beta}^+(s)$ and tight to states from $d_{\beta}^-(s)$), such that $D(\pi, \beta)(s_1)$ is large and $D(\pi, \beta)(s_2)$ is small. $C_{\text{diff}}^\pi$ would be relatively low, as the numerator, the policy divergence $D(\pi, \beta)(s)$, increases monotonically in relation to the denominator, the coverage of the dataset $d_\beta(s)$. Small $C_{\text {diff }}^\pi$ indicates that the policy $\pi$ stays close to the behavior policy $\beta$ in $d_{\beta}^-(s)$ while deviates significantly in $d_{\beta}^+(s)$, consisting with our didactic example and conclusion in \Cref{subsec: state-agnostic}.

Given the definition of differential concentrability under the imbalanced state coverage, we follow \citet{singh2022offline} and use it to bound the policy improvement over $\pi$ w.r.t. $\beta$ in the safe policy improvement framework \cite{laroche2019safe}. We then show that large $C_{\text {diff }}^\pi$ is hard to guarantee the policy improvement with a large margin, under the distributional constraints methods in \Cref{eqn: policy constrain}:

\begin{theorem}[Limited policy improvement via distributional constraints.] W.h.p. $\ge 1-\delta$, for any prescribed level of safety $\zeta$, the maximum possible policy improvement over choices of $\alpha$, $J\left(\pi\right)-J\left(\beta\right) \leq \zeta^{+}$, where $\zeta^{+}$ is given by: 
\begin{equation}
\zeta^{+}:=\max _\alpha \quad h^*(\alpha) \cdot \frac{1}{(1-\gamma)^2}
\label{eq: policy improvement}
\end{equation}
\begin{equation*}
\text { s.t. } \frac{c_1}{(1-\gamma)^2} \frac{\sqrt{C_{\text {diff }}^{\pi_\alpha}}}{|\mathcal{D}|}-\frac{\alpha}{1-\gamma} \mathbb{E}_{s \sim d_\pi(s) }\left[D\left(\pi, \beta\right)(s)\right] \leq \zeta
\end{equation*}
where $h^*$ is a monotonically decreasing function of $\alpha$, and $h(0)=\mathcal{O}(1)$.
\label{theo: policy improvement}
\end{theorem}

\Cref{theo: policy improvement} expresses the essential trade-off between distributional constraints and safe policy improvement. The LHS of the constraint in \Cref{eq: policy improvement} is a monotonically increasing function of $C_{\text {diff }}^\pi$. When the value of $C_{\text {diff }}^\pi$ is low, a relatively small $\alpha$ can lead to a significant improvement in performance, as measured by $h^*(\alpha)$ in \Cref{eq: policy improvement}, while also ensuring compliance with the $\zeta$-safety condition. However, when the dataset is exposed to imbalance (i.e., $C_{\text {diff}}^\pi$ is large) in the current state-agnostic constrains framework, satisfying the $\zeta$-safety condition requires a larger $\alpha$, which obtains smaller maximum possible improvement $h^*(\alpha)$. When the degree of imbalance $\eta$ in \Cref{eqn: exponential distribution} approaches extremes, the improvement $h^*(\alpha)$ over the learned policy $\pi$ is close to zero with high likelihood, i.e., the learned policy fails to improve over the behavior policy $\beta$. We remark that smaller $C_{\text {diff }}^\pi$ is essential for the $\zeta$-safety condition and hence a smaller $\alpha$ for the improvement $h^*(\alpha)$ with a large margin. The proofs of \Cref{theo: policy improvement} can be found in \citep{singh2022offline}.

\section{Retrieval Augmented Offline RL}
As discussed in~\Cref{sec3: current offline RL methods fail} with empirical and theoretical analyses, current state-of-the-art offline RL algorithms can hardly solve imbalanced offline RL datasets.
In Natural Language Processing (NLP) and computer vision (CV) communities, there is a line of works~\citep{kandpal2022large, long2022retrieval, wu2008information} that leverages external information or knowledge to alleviate the heavy-tail or long-tail problems. 
Actually, our designed imbalanced dataset in offline RL is similar to these studies, which are following Zipf's law. As a preliminary study, we try to introduce non-parametric retrieval-augmented methods to enhance standard offline RL algorithms by combining retrieved relevant states from external resources.

\noindent\textbf{Auxiliary dataset preparation.}
Firstly, we prepare external resources as our auxiliary dataset $\mathcal{D}_{\text{aux}}$ for the retrieval process. The auxiliary dataset contains a large body of knowledge or experience, for example, trajectories from previous experiences of the current learned policy or other agents \citep{goyal2022retrieval, humphreys2022large}.
In our setups, we directly adopt the state information in the same and relevant environments as auxiliary dataset $\mathcal{D}_{\text{aux}}$.



\noindent\textbf{Retrieval process.}
Secondly, we retrieve the nearest states with useful information from auxiliary dataset $\mathcal{D}_{\text{aux}}$.
Formally, given the query state $s_{\text{ori}}^{i} \in \mathcal{D}$ and the external state $s_{\text{aux}}^{i} \in \mathcal{D}_\text{aux}$ from the auxiliary dataset, we encode all these states to representations as static indexes. For high dimensional states, 
the representations of both query state and external states can be denoted as $\text{Enc}(s_{\text{ori}}^{i})$ and $\text{Enc}(s_{\text{aux}}^{i})$. To mitigate additional computation burden and eliminate effects from other architecture, we evaluate our method on Mujoco \citep{todorov2012mujoco} tasks with low dimensional states. 
Then, after building the indexes, we compute the similarity score between these indexes based on Maximum Inner Product Search (MIPS) or other distance functions, such as Euclidean distance.
Taking the dense inner product as an example, the detailed similarity score can be computed as follows:
\begin{equation}
    \text{Sim}(s_{\text{ori}}^{i}, s_{\text{aux}}^{i}) = \frac{\exp(s_{\text{ori}}^{i}\cdot  s_{\text{aux}}^{i})}{\sum_{j=1}^{\left|\mathcal{D}_{\text{aux}}\right|}\exp(s_{\text{ori}}^{i}\cdot  s_{\text{aux}}^{j})}
\end{equation}

According to the similarity scores $\text{Sim}(s_{\text{ori}}^{i}, s_{\text{aux}}^{i})$ between the query state $s_{\text{ori}}^{i}$ and external states $s_{\text{aux}}^{i}$, our retrieval process selects the top-$k$ nearest retrieved states $\{s_{\text{ret}}^1, s_{\text{ret}}^2, \cdots, s_{\text{ret}}^k\}$ to augment our agent, which will further introduce in the next part.
Following \citet{humphreys2022large}, we use $k=10$ and adopt the \texttt{SCaNN} architecture~\citep{guo2020accelerating} as the implementation to efficiently complete the approximate vector similarity search.

\noindent\textbf{Retrieval-augmented agent.} 
Finally, we hope to leverage these retrieved states $\{s_{\text{ret}}^1, s_{\text{ret}}^2, \cdots, s_{\text{ret}}^k\}$ for our agent to alleviate the heavy-tail problem reflected in our exploration-challenging dataset. 
With query state $s_{\text{ori}}^i$, our agent further exploits the retrieved states by a straightforward and no-parametric approach.
Considering that directly concatenating all retrieved states to the query state will lose the main information of the query state, we concatenate the original state $s_{\text{ori}}^i$ with averaged retrieved states to produce the final state $s_{\text{final}}$:
\begin{equation}
    s_{\text{final}}=\left[s_{\text{ori}}^i\oplus\frac{\sum^{k}_{j=1}s_{\text{ret}}^j}{N}\right]
\end{equation}
Where $\oplus$ is the concatenate operator.
We then feed the final state $s_{\text{final}}$ to the policy network and value network, then further adopt the popular value-based offline RL method, e.g., CQL, to train and evaluate. It is worth noting that the only difference between the offline RL method and our retrieval-augmented offline RL method is that we introduce additional state information via $s_{\text{final}}$ instead of the only state from the current transition $s_{\text{ori}}^i$. We summarize the pseudo-code of RB-CQL in \Cref{appendix:pseudo code}.

\section{Experiments}

\begin{figure}[t]
\centering
\includegraphics[width=0.5\linewidth]{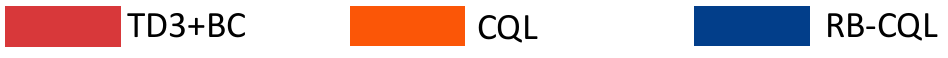} 
\subfloat[AntMaze-medium]{\includegraphics[width=1\linewidth]{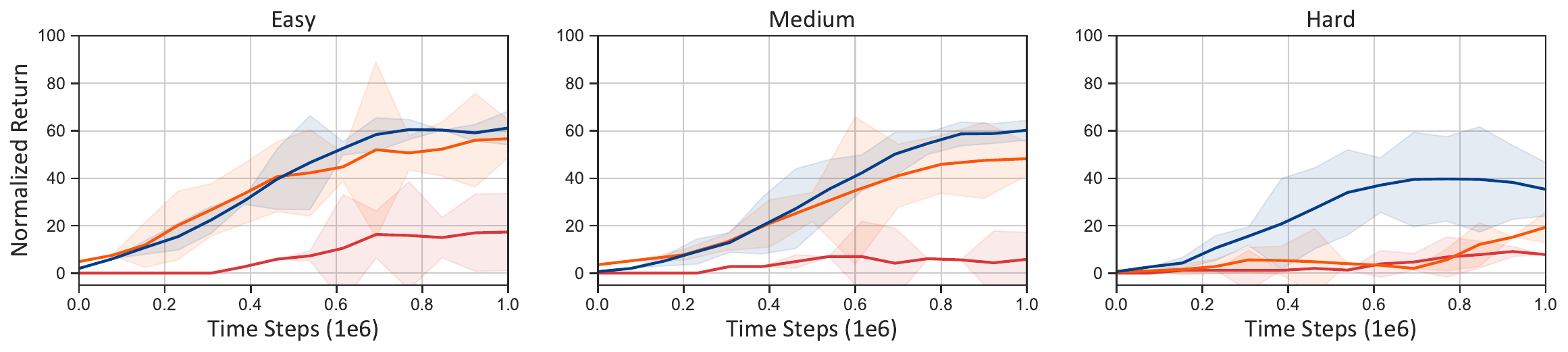}}
\vspace{-3mm}
\subfloat[AntMaze-large]{\includegraphics[width=1\linewidth]{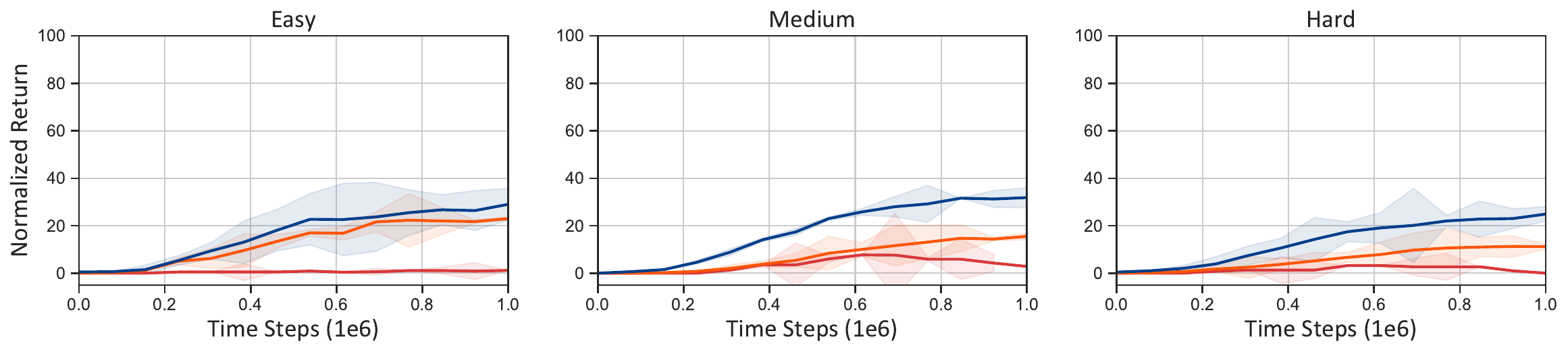}}
\vspace{-3mm}
\subfloat[Mujoco-hopper]{\includegraphics[width=1\linewidth]{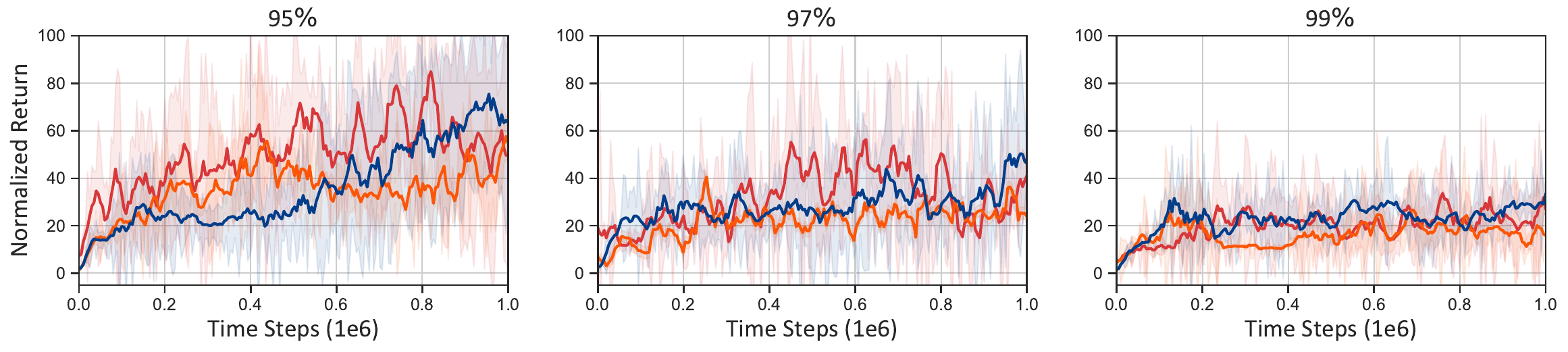}}
\vspace{-3mm}
\subfloat[Mujoco-walker2d]{\includegraphics[width=1\linewidth]{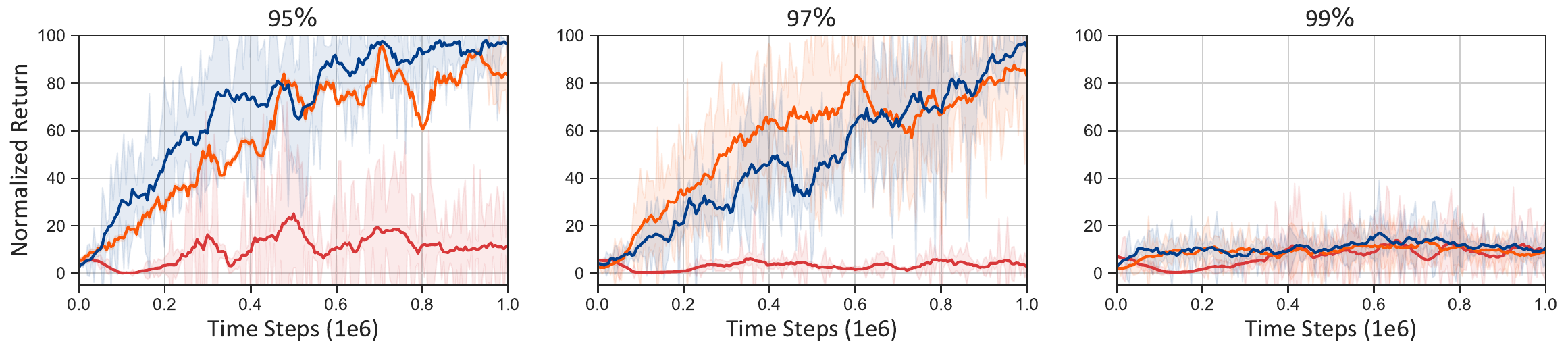}}
\vspace{-2mm}
\caption{Average normalized scores of RB-CQL against other baselines over the whole training process. Regarding the final checkpoint, RB-CQL reaches the best performance in 10 out of 12 tasks, where an increasing margin with the increasing imbalance in AntMaze.}
\label{results:benchmark}
\vspace{-4mm}
\end{figure}

\begin{figure*}[t]
    \centering
    \subfloat[Query state]{
        \includegraphics[width=0.15\textwidth]{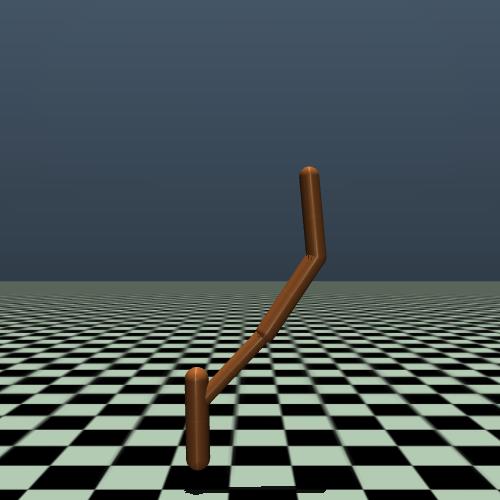}
    }
    \hspace{5mm}
    \subfloat[Retrieved states]{
        \includegraphics[width=0.15\textwidth]{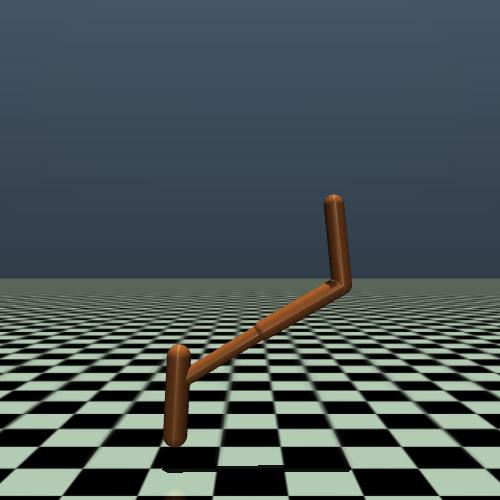}
        \includegraphics[width=0.15\textwidth]{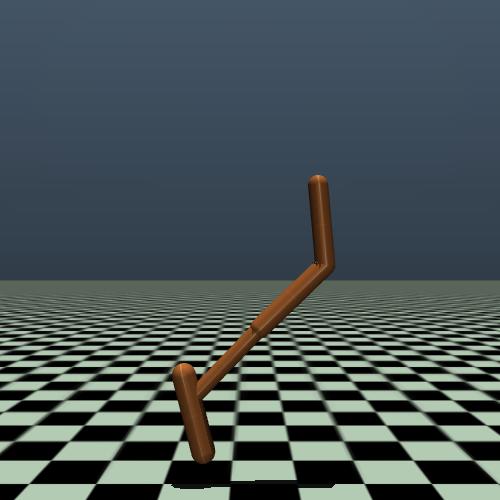}
        \includegraphics[width=0.15\textwidth]{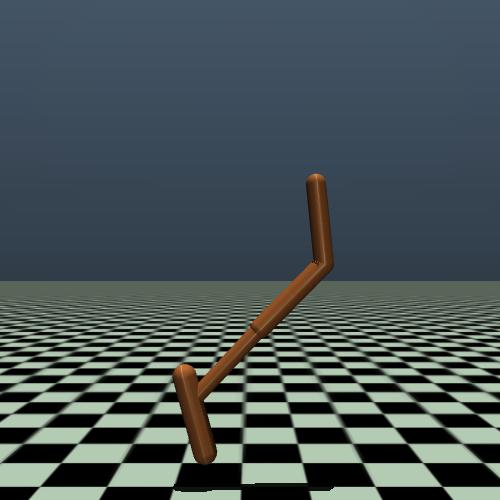}
        \includegraphics[width=0.15\textwidth]{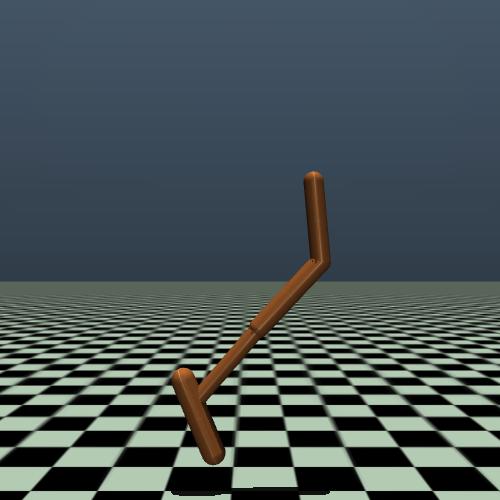}
        \includegraphics[width=0.15\textwidth]{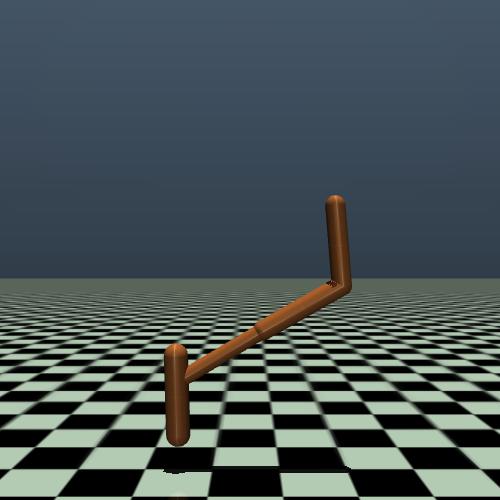}
    }
    \quad
    \subfloat[Query state]{
        \includegraphics[width=0.15\textwidth]{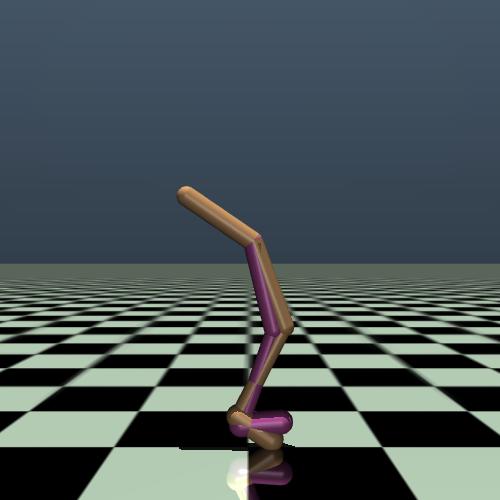}
    }\hspace{5mm}
    \subfloat[Retrieved states]{
        \includegraphics[width=0.15\textwidth]{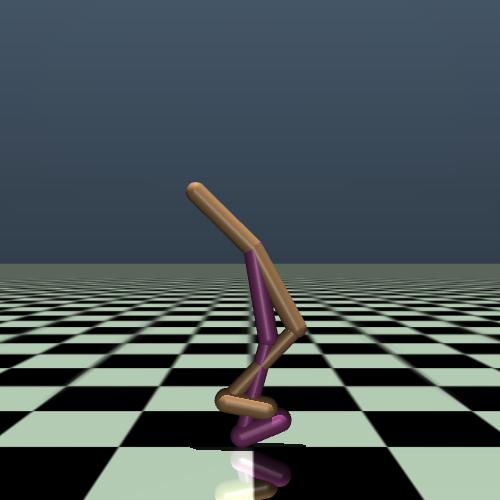}
        \includegraphics[width=0.15\textwidth]{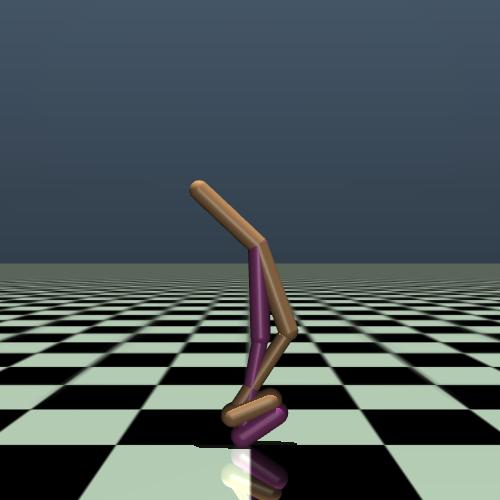}
        \includegraphics[width=0.15\textwidth]{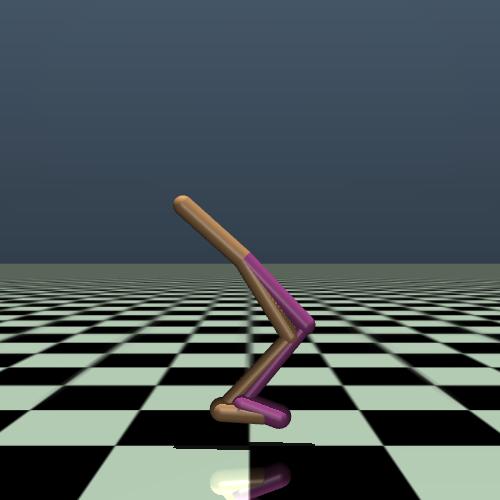}
        \includegraphics[width=0.15\textwidth]{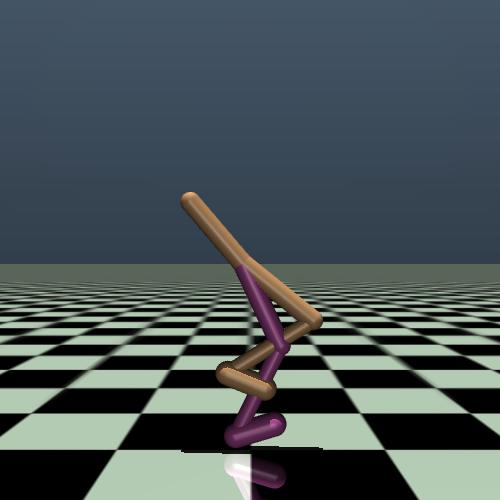}
        \includegraphics[width=0.15\textwidth]{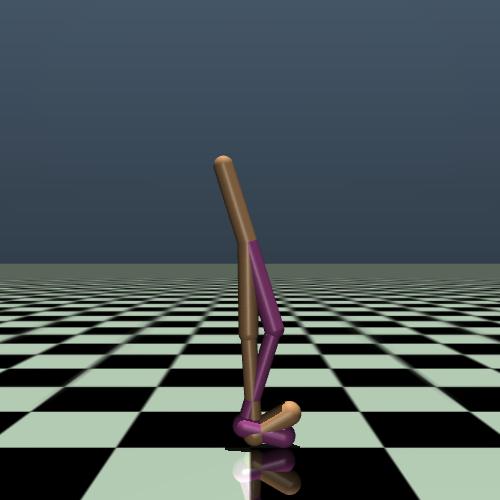}
    }

    \caption{Visualisation of N=5 retrieved from RB-CQL for Mujoco locomotion tasks (first row for Hopper and second row for Walker2d.)}
    \label{fig:retrieved_example}
\end{figure*}
\textbf{Imbalanced datasets.} In order to evaluate the effectiveness of Retrieval-based CQL (RB-CQL), we first establish four new tasks from two tasks with imbalanced datasets in the variant of the standard benchmark D4RL \citep{fu2020d4rl}. Recall in \Cref{sec3: current offline RL methods fail}, the distribution of imbalanced datasets follows Zipf's law, and the magnitude of its coverage is negatively related to the degree of policies (i.e., sufficient to insufficient coverage is characterized by mixture and expert policies). For AntMaze tasks, we use a goal-reaching controller with a fixed start distribution (same as the evaluation process) and a goal distribution varying in the whole state space, where only rare goal distributions are identical to the evaluation process. It represents that the dataset is marked by imbalanced state coverage, where rare expert trajectories could reach the goal state in terms of evaluation. 
For Mujoco tasks, we simply combine the random and expert dataset within a few expert datasets. The increasing $\eta$ in \Cref{eqn: exponential distribution} denotes the increasing imbalance over those datasets: for AntMaze navigation tasks in medium and large mazes, smaller near-optimal trajectories are corresponding from \texttt{Easy} to \texttt{Hard+} dataset; for Mujoco locomotion tasks, smaller near-optimal trajectories are corresponding to random dataset ratio from $95\%$ to $99\%$. Note that AntMaze tasks are more challenging than Mujoco locomotion tasks due to the sparse reward setting and the heavy requirement for exploration.

We then compare RB-CQL with TD3+BC \cite{fujimoto2021minimalist} and CQL \cite{kumar2020conservative}, which are two state-of-the-art methods and enforce policy support constraints. For a fair comparison, we sweep important hyperparameters of each algorithm. The experimental results for these tasks are reported in \Cref{results:benchmark}. The shaded area in the plots represents the standard deviation of the results. Detail implementation and experimental details are provided in \Cref{appendix: exp details} and the performance profiles \citep{agarwal2021deep} based on score distribution are shown in \Cref{fig:performance proﬁles}.

For each AntMaze task with increasing imbalance (i.e., \texttt{Easy} to \texttt{Hard}), our approach, RB-CQL, is robust to the imbalance and maintains its performance to a certain degree, especially from \texttt{Easy} to \texttt{Medium} without any performance drop. However, the performance of other methods is vulnerable to the imbalance in the dataset, where they have a sharp decrease (e.g., CQL decreases from 60 to 20, and TD+BC decreases from 20 to 0 in AntMaze-medium). The performance gap between RB-CQL and CQL grows with the increasing imbalance (i.e., from \texttt{Easy} to \texttt{Hard}), which illustrates the effectiveness of the retrieval augmentation on CQL. All algorithms, including RB-CQL, fail in the extremely hard AntMaze task \texttt{Hard+}, shown in \Cref{appendix:add experiences}. In Mujoco locomotion tasks, known for easier tasks compared to AntMaze, RB-CQL outperforms other methods with $95\%$ and $97\%$ random dataset ratio. Similarly, in the heavily imbalanced dataset of the {Hard+} AntMaze task, there was limited improvement observed with the use of the RB-CQL method compared to other methods. 

We also visualized what we retrieved from our retrieval process, shown in \Cref{fig:retrieved_example}. From the locomotion perspective, the posture is not exactly the same between the query state and retrieved states, but they are similar from the macroscopic point of view. It boosts the generalization ability by the augmentation from the auxiliary dataset.

\section{Related Work}
Ofﬂine RL \citep{ernst2005tree, riedmiller2005neural, lange2012batch, levine2020offline} is interested in the problem of extracting a policy from a fixed dataset, without further interacting with the environment. It provides the promise to many real-world applications where data collection is expensive or dangerous such as healthcare \citep{tang2021model}, autonomous driving \citep{kalashnikov2021mt}, industrial control \citep{zhan2022deepthermal}, and task-oriented dialog systems \citep{verma2022chai}. The task of offline RL, learning a policy to perform better than the behavior policy in the dataset, often leads to the distributional shift between the learned policy and the behavior policy \citep{fujimoto2019off, levine2020offline}, which can cause erroneous extrapolation error. 

\textbf{Offline RL without dataset consideration.} 
To alleviate the above problem, previous methods enforce pessimism in actor updates \citep{wu2019behavior, fujimoto2019off, kumar2019Stabilizing, fujimoto2021minimalist, xu2021offline, zhou2021plas} or critic updates \citep{liu2020provably, kumar2020conservative, bai2022pessimistic, li2022data}, whose learned policy would query the Q-functions of out-of-distribution actions to outperform the behavior policy, implicitly or explicitly minimizing the divergence over those two policies to mitigate the distributional shift problem. Another branch of methods learns an optimal policy by using only in-sample transitions, also called in-sample learning, bypassing querying the values of unseen actions. This can be achieved by vanilla advantage-weighted regression \citep{peng2019advantage, nair2020awac}, the extreme approximation of advantage-weighted regression \citep{chen2020bail, kostrikov2021iql, xu2023offline, brandfonbrener2021quantile}, conditioning on some future information or target information \citep{kumar2019reward, chen2021decision, emmons2021rvs, feng2022curriculum}, or find the best next states in datasets \citep{xu2022policy}. Those methods have achieved significant success in standard benchmarks \citep{gulcehre2020rl, fu2020d4rl, qin2021neorl}, however, they have not adequately considered the properties of the dataset, despite it being the primary source to extract policies. In this study, we shift our focus towards offline RL for real-world applications by taking into account the distribution of real-world datasets (i.e., imbalanced datasets), and examining the limitations of current algorithms in such scenarios.

\textbf{Offline RL with dataset consideration.} 
The prevalent use of benchmarks \citep{gulcehre2020rl, fu2020d4rl, qin2021neorl} in current research has led to a neglect of the importance of the properties of the dataset in the development of models. Previous benchmarks have often assumed that datasets are not imbalanced, with rare but important trajectories, despite the fact that studies such as \citep{qin2021neorl} provides dataset from real-world scenarios and recognizes that the exploration challenge is a fundamental challenge in offline RL. Several prior works have conducted extensive empirical analyses to establish various characteristics \citep{swazinna2021measuring, schweighofer2021understanding} regarding the impact of dataset properties on performance. Despite these efforts, previous studies have primarily focused on standard benchmarks, without taking into account real-world datasets and have not adequately addressed the practical challenges posed by imbalanced datasets. While policies fail to adequately learn under such multi-task or lifelong learning settings with imbalanced datasets from different environments \citep{zhou2022forgetting, chan2022zipfian}, the single-task setting has been hardly investigated. Instead of allowing the agent to further actively interact with the environment to collect task-specific or task-agnostic datasets to overcome the imbalance or exploration challenge \citep{riedmiller2022collect, yarats2022don, lambert2022challenges}, we are interested in the setting that additional interaction is forbidden but other large-scale offline datasets are available. 

\textbf{Retrieval in RL.}
Retrieval-augmented methods have recently been widely used to leverage a large body of external information to support downstream tasks in natural language processing~\citep{chen-etal-2017-reading,karpukhin-etal-2020-dense,izacard-grave-2021-leveraging} or computer vision fields~\citep{gur2021cross,zhu2020caesar}.
Recently, the RL community introduces the retrieval system to integrate experience with new information in a straightforward way to advise the current decision: mitigate the expensive computation and overcome the capacity limitation of models \citep{goyal2022retrieval, humphreys2022large}. With the same purpose of utilizing related experiences to facilitate decisions, \citet{paischer2022history} use a frozen pre-trained language transformer to embed history information and improve sample efficiency in the partially observable Markov decision process.

The methods mostly similar to ours, RB-CQL, are \citet{goyal2022retrieval, humphreys2022large}, where they both augment a standard reinforcement learning agent with a retrieval process. However, the motivation is different. Our algorithm is designed to overcome the difficulties of the imbalanced datasets in the offline RL setting, while they are aiming to reduce the computation burden and integrate the previous experience into the model without additional gradient updates. Besides, the retrieval architecture of those two algorithms is parametric, which needs other computation sources, but ours is non-parametric. 
\section{Conclusions and Future Work}

In this work, we investigate the challenges posed by imbalanced datasets in offline reinforcement learning, which mirror the distribution commonly found in real-world scenarios. We first characterize the imbalanced offline RL dataset with an imbalanced state coverage, varying policies ranging from a mixture of behaviors to expert ones from sufficient to poor state coverage. Empirically and theoretically, we show that the current offline RL method, CQL, is flawed in extracting high-quality policies under such datasets. 
To overcome the challenges in imbalanced datasets, We further propose a new method, retrieval-based CQL (RB-CQL), by the augmentation of CQL with a retrieval process, which directly incorporates related experiences from a larger source dataset into training. We evaluate our method on a range of tasks with varying levels of imbalance using the variant of D4RL and show that our method is robust to the imbalance and outperforms other baselines. Limitations of our work are that the retrieval process requires huge computation sources from the CPU and the lack of study on high dimensional inputs, which requires the embedding networks, to exclude other influences. We hope this work will inspire further research in the field of offline RL, particularly utilizing practical and imbalanced datasets, and serve as a foundation for large-scale offline RL applications. 

%
\bibliography{main.bbl}
\bibliographystyle{icml2021}

\clearpage

\onecolumn
\icmltitle{Offline Reinforcement Learning with Imbalanced Datasets: \newline Supplementary Material}
\icmltitlerunning{Offline Reinforcement Learning with Imbalanced Datasets: Supplementary Material}
\appendix

\section{Pseudo Code} \label{appendix:pseudo code}
\begin{algorithm}[htb!]
\caption{Retrieval-Augmented Offline Reinforcement Learning}
\label{alg1}
	\begin{algorithmic}[1] 
		\Require Current training offline dataset $\mathcal{D}$, retrieval auxiliary dataset $\mathcal{D}_\text{aux}$, training timestep $T$.
		\State \textbf{Initialize} $R_{\eta}$, $\pi_{\theta}$
            \State {\color{blue}{// Training}}
		\For{$i=1,2,\ldots,T$} 
                \State Sample external states $s_\text{aux} \sim \mathcal{D}_\text{aux}$
                \State Sample transitions $(s, r, a, s') \sim \mathcal{D}$ 
                \State Select top-$k$ retrieved states $s_\text{ret}$ and $s'_\text{ret}$ via $\text{Sim}(s, s_{\text{aux}}) = \frac{\exp(s\cdot s_{\text{aux}})}{\sum_{j=1}^{\left|\mathcal{D}_{\text{aux}}\right|}\exp(s\cdot s_{\text{aux}}^j)}$
                \State Get concatenated states $\widetilde{s} = \left[s \oplus s_\text{ret}\right]$, $\widetilde{s'} = \left[s \oplus s'_\text{ret}\right]$ 
                \State Use $(\widetilde{s}, r, a, \widetilde{s}')$ for downstream offline RL training
            \EndFor   
            \State \textbf{Return} policy $\pi_{\theta}$
            
            \State {\color{blue}{// Evaluation}}
            \State Get initial state $s$, set $d$ as False
            \While{not $d$}:
                \State Select top-$k$ retrieved states $s_\text{ret}$ via the same $\text{Sim}(s, s_\text{aux})$
                \State Get concatenated states $\widetilde{s} = \left[s \oplus s_\text{ret}\right]$
                \State Get action  $a=\pi_{\theta}(\widetilde{s})$
                \State Roll out $a$ and get $(s', r, d)$
                \State Set $s=s'$
            \EndWhile
	\end{algorithmic}
\end{algorithm}

\section{Additional Experiences} \label{appendix:add experiences}

\begin{figure}[h]
    \centering
        \includegraphics[width=0.23\textwidth]{fig/heavy_tail/heavy_tail_with_per_0.pdf}
        \includegraphics[width=0.23\textwidth]{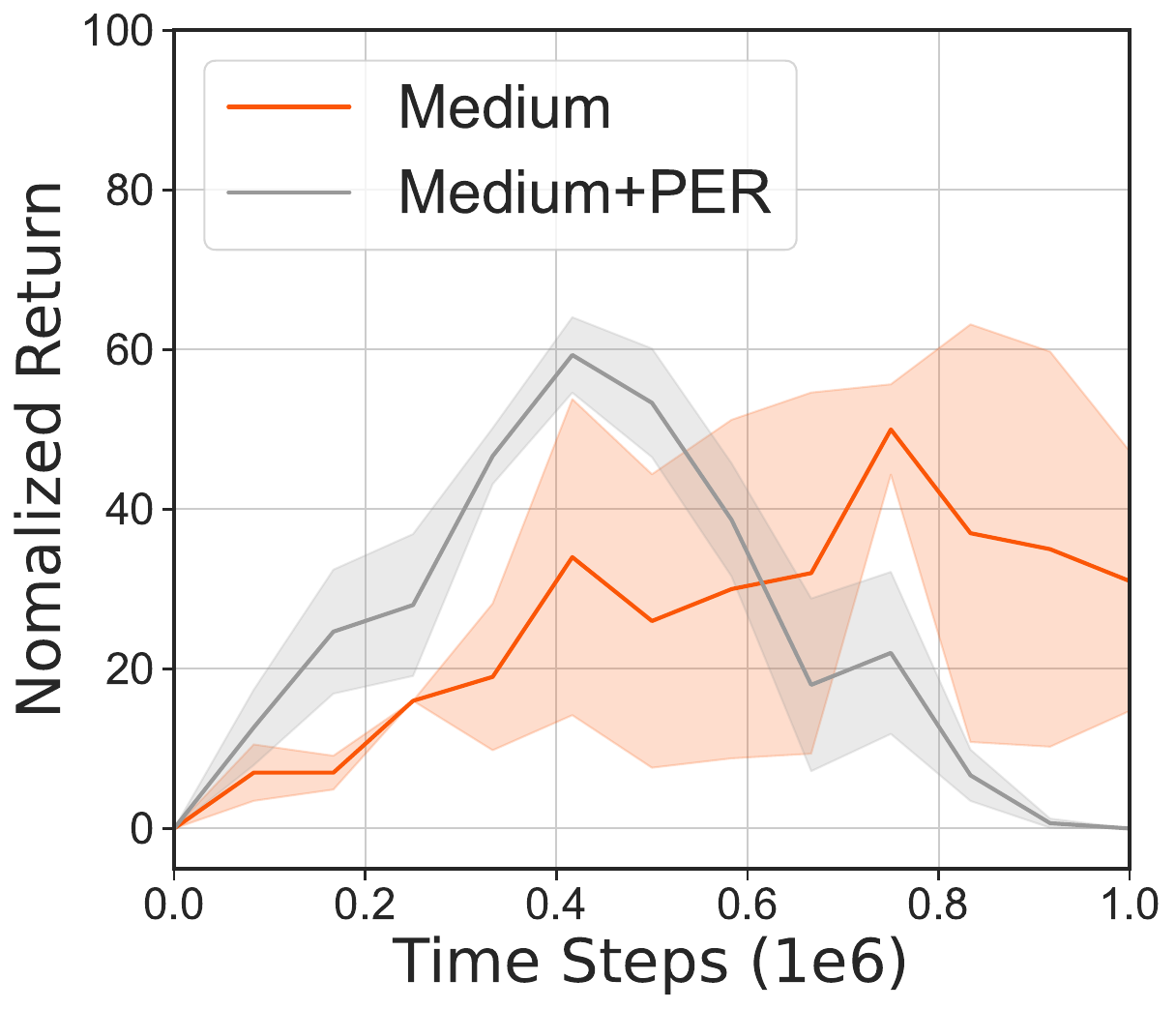}
        \includegraphics[width=0.23\textwidth]{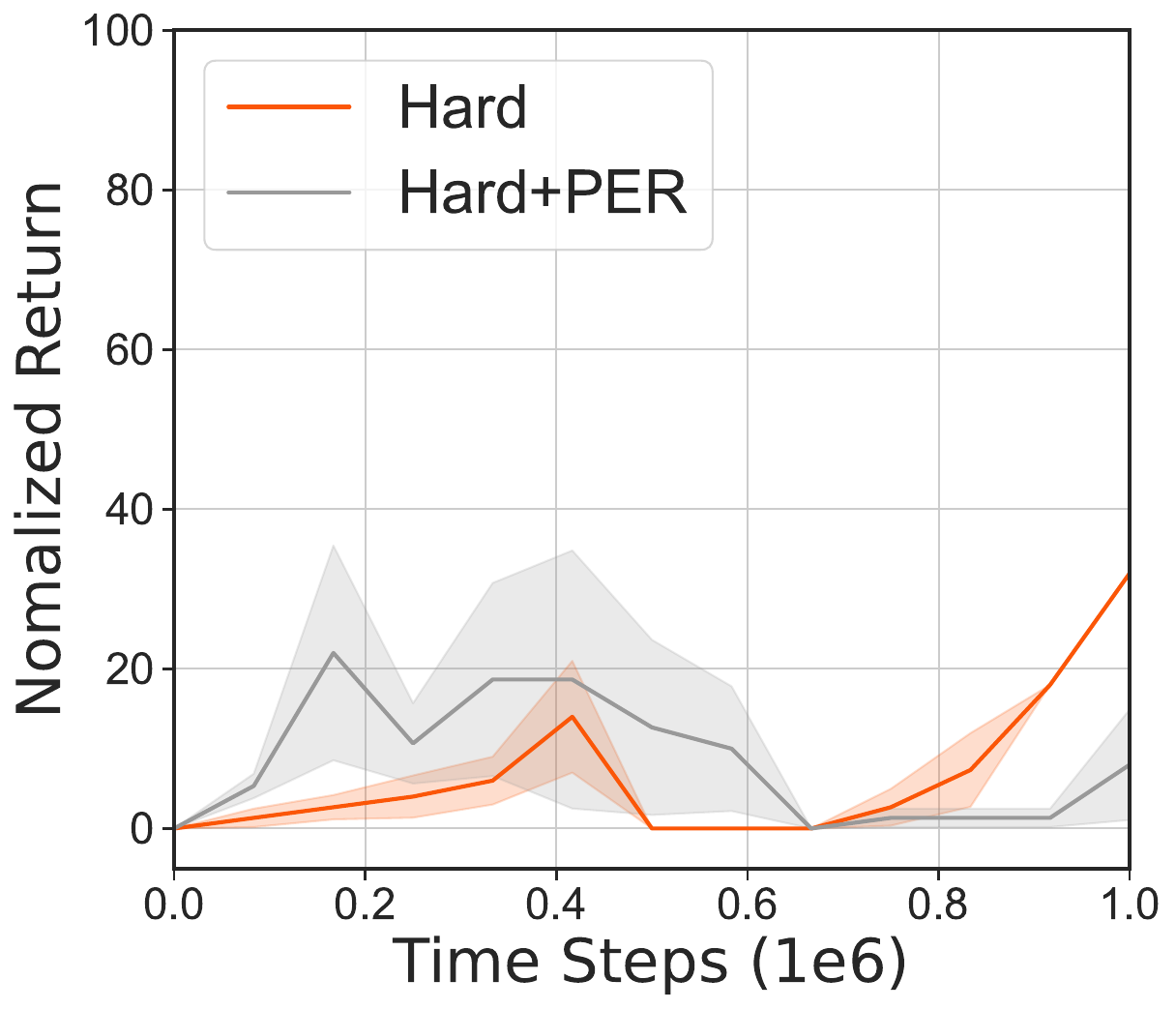}
        \includegraphics[width=0.23\textwidth]{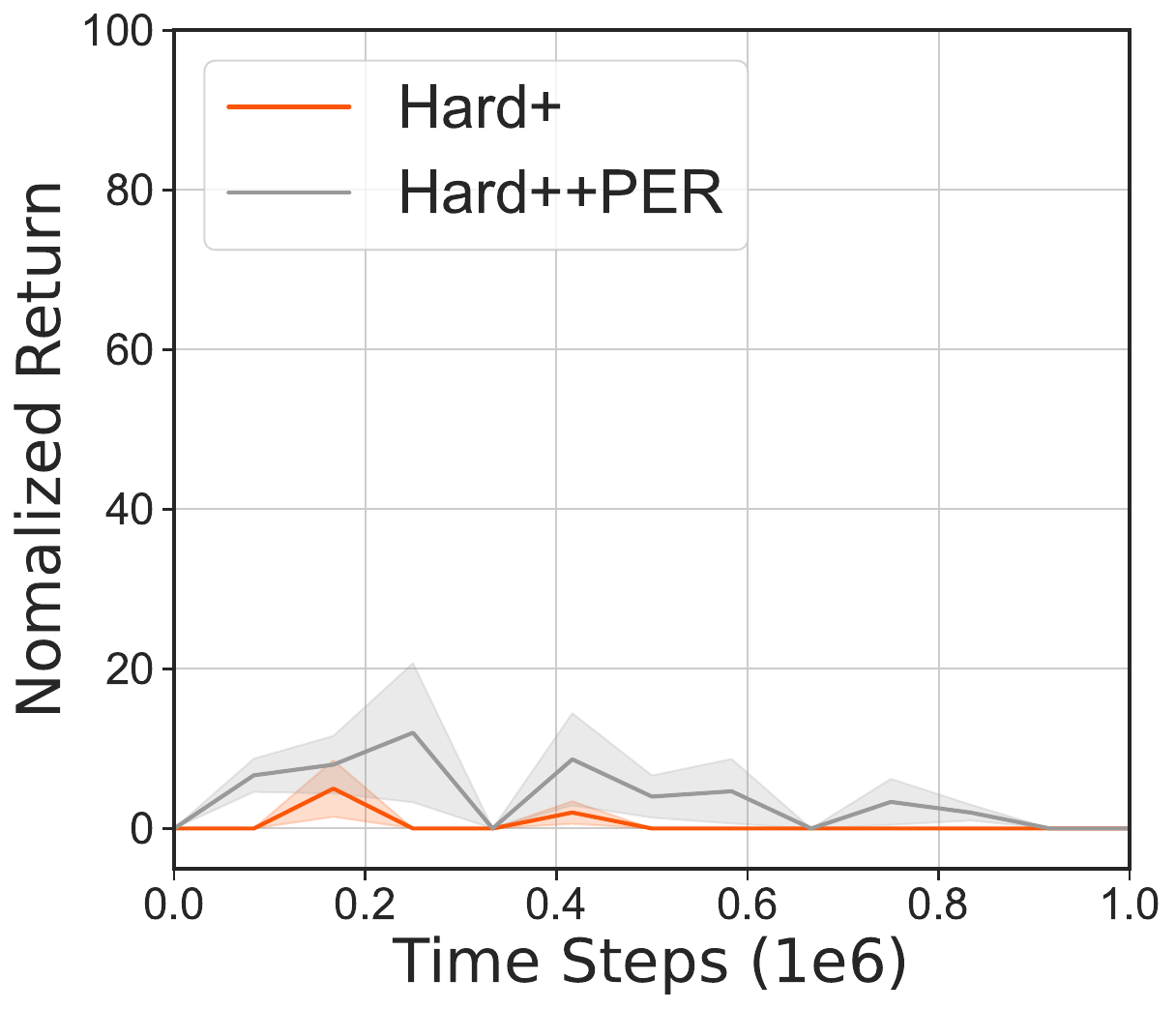}   
    \caption{The performance of CQL and CQL with PER on \texttt{Antmaze-medium} task on easy to hard+ level imbalance. Except for the hard+ task, CQL with the augmentation of PER worsens the final performance, compared with CQL.)}
    \label{fig:additional exp on per}
\end{figure}

\begin{figure}[h]
    \centering
    \includegraphics[width=0.5\textwidth]{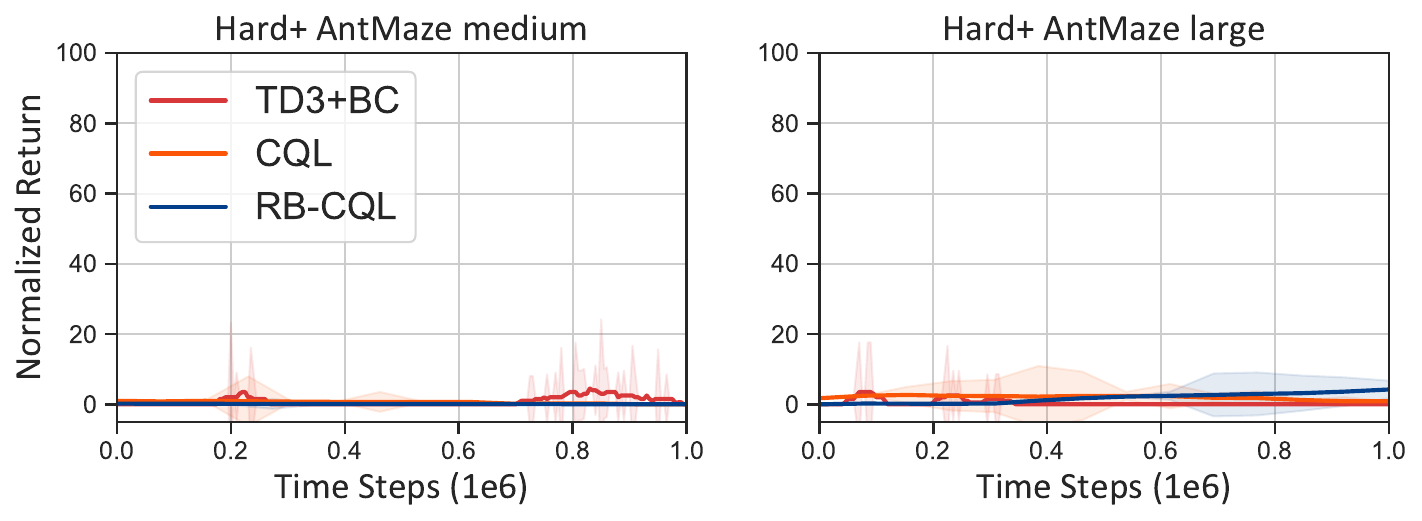}
    \caption{Average normalized scores of RB-CQL against other baselines over the whole training process in the extremely hard task, i.e., hard+.}
    \label{fig:my_label}
\end{figure}

\begin{figure}
    \centering
    \includegraphics[width=0.45\textwidth]{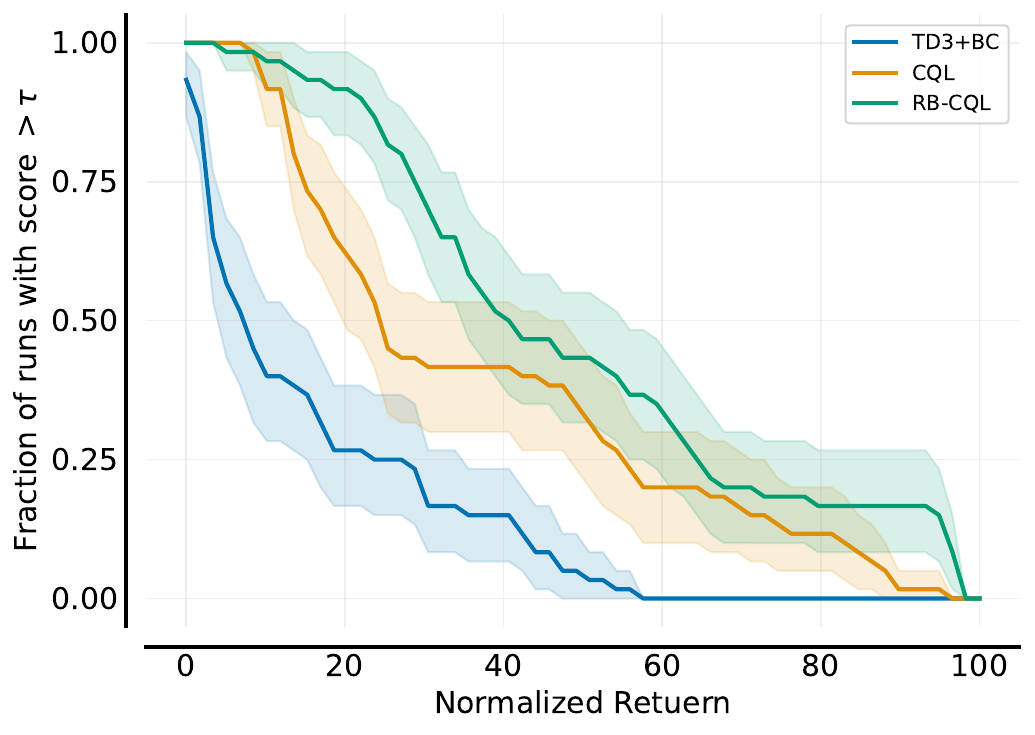}
    \includegraphics[width=0.45\textwidth]{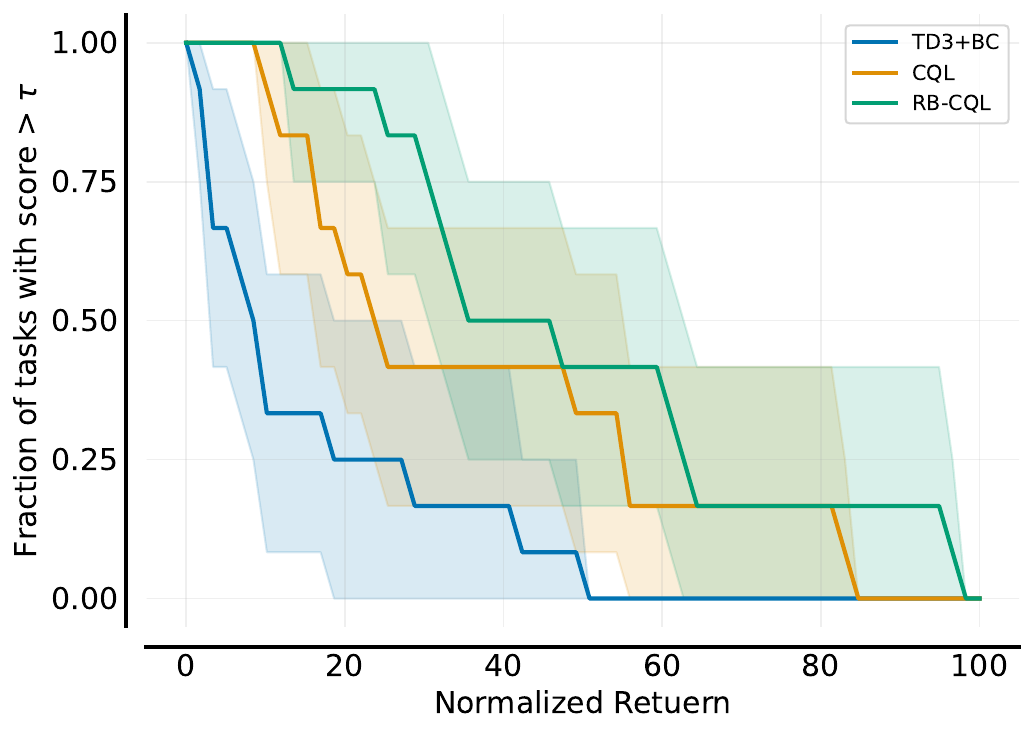}
    \caption{Performance proﬁles on the imbalanced dataset based on score distributions over five runs and an average of five runs (tasks) on every task. Shaded regions show pointwise 95\% conﬁdence bands based on percentile bootstrap with stratiﬁed sampling.}
    \label{fig:performance proﬁles}
\end{figure}

\section{Experiment Details} \label{appendix: exp details}

\label{appendix: exp}
\textbf{The variant of D4RL AntMaze datasets.} The dataset we introduced in this work is all imbalanced with the heavy-tail property, featuring imbalanced coverage and rare but important trajectories. In AntMaze tasks, we introduce four new AntMaze datasets for the medium and large mazes from D4RL \cite{fu2020d4rl}. For each type of maze, the imbalance increases from \texttt{Easy} to \texttt{Hard+}, characterizing an optimal decreasing trajectory that can reach the target state distribution in evaluation. In detail, the start state distribution is fixed and aligned with the evaluation, whereas the goal state distribution can be varying and misaligned with the evaluation. In each episode, the goal state distribution can hardly be aligned with the goal state distribution in evaluation. Given an observation, a goal state from the environment, and optimal action from a pre-trained optimal goal-reaching policy, the agent executes the correct action given by the optimal policy. That simulates the imbalanced dataset we have introduced in our work, where the state coverage is heavy-tail and the rare trajectories near the target state are optimal in terms of evaluation.  On the other hand, on Mujoco locomotion tasks, we simply combine the random and expert datasets with large ratios approaching 100$\%$ (i.e., 95$\%$, 97$\%$, and 99$\%$). It also stimulates the imbalanced dataset we have described because only expert policies can reach certain states and thus gain significant performance, while random policies reach those states with low likelihood. The total transitions in every dataset are 1M. To give a more intuitive representation of the imbalance over the whole state space, we calculate the number of transitions that are in the start distribution grid, a medium grid at the center of the maze, and the target goal state distribution grid, shown in \Cref{tab:imbalced data antmaze}. We also present the total transitions and expert transitions in imbalanced Mujoco locomotion tasks, shown in \Cref{tab:imbalanced data mujoco}.

\begin{table*}[htbp!]
\centering
\normalsize
\caption{Details of the imbalanced dataset in AntMaze tasks.}
\label{tab:imbalced data antmaze}
\begin{tabular}{cl|rrrr} 
\toprule
\multicolumn{2}{l|}{Dataset (AntMaze)} & \multicolumn{1}{c}{Total tansitions} & \multicolumn{1}{c}{Start} & \multicolumn{1}{c}{Meidum} & \multicolumn{1}{c}{Target}  \\ 
\hline
\multirow{5}{*}{medium} & Vanilla      & 100,000                              & 95, 000                   & 51,000                     & 6,000                       \\
                        & Easy         & 100,000                              & 104, 000                  & 45,000                     & 800                         \\
                        & Medium       & 100,000                              & 83, 000                   & 45,000                     & 300                         \\
                        & Hard         & 100,000                              & 84, 000                   & 63,000                     & 90                          \\
                        & Hard+        & 100,000                              & 90, 000                   & 64,000                     & 21                          \\ 
\hline
\multirow{5}{*}{large}  & Vanilla      & 100,000                              & 105, 000                  & 45,000                     & 8,000                       \\
                        & Easy         & 100,000                              & 100, 000                  & 41,000                     & 4,000                       \\
                        & Medium       & 100,000                              & 87, 000                   & 54,000                     & 800                         \\
                        & Hard         & 100,000                              & 92, 000                   & 62,000                     & 350                         \\
                        & Hard+        & 100,000                              & 86, 000                   & 64,000                     & 100                         \\
\bottomrule
\end{tabular}
\end{table*}

\begin{table*}[htbp!] 
\centering
\normalsize
\caption{Details of the imbalanced dataset in Mujoco locomotion tasks.}
\label{tab:imbalanced data mujoco}
\begin{tabular}{c|crrr}
\toprule
Env                          & Random ratio  & \multicolumn{1}{c}{Total transitions} & \multicolumn{1}{c}{Expert transitions} & \multicolumn{1}{c}{Random transitions}  \\ 
\hline
\multirow{3}{*}{Hopper}    & 95\%               & 1,000,000                               & 50,000                                  & 950,000                                  \\
                             & 97\%               & 1,000,000                               & 30,000                                  & 970,000                                  \\
                             & 99\%              & 1,000,000                               & 10,000                                 & 990,000                                  \\
\hline
\multirow{3}{*}{Walker2d} & 95\%               & 1,000,000                               & 50,000                                  & 950,000                                  \\
                             & 97\%               & 1,000,000                               & 30,000                                  & 970,000                                  \\
                             & 99\%              & 1,000,000                               & 10,000                                 & 990,000                                  \\
\bottomrule                            
\end{tabular}
\end{table*}

For MuJoCo locomotion tasks, we average mean returns over 10 evaluations every 5000 training steps. For AntMaze tasks, we average over 100 evaluations every 0.1M training steps. They are both taken from 5 random seeds. For the source code of our experiences, we re-run TD+BC from author-provided source code \href{https://github.com/sfujim/TD3_BC}{https://github.com/sfujim/TD3\_BC}. Due to the introduction of new datasets and for a fair comparison, we sweep the hyper-parameter $\alpha$ in TD3+BC in $[0.1, 0.2, 0.5, 0.7, 1]$ and choose the best performance in our results. For CQL and our method, we use the open-source JAX \citep{jax2018github} version code: \href{https://github.com/young-geng/JaxCQL}{https://github.com/young-geng/JaxCQL}. Same as we do in TD3+BC, we sweep the hyper-parameter $\alpha$ in CQL in $[5, 10, 20, 50]$ and choose the best performance in our results. Note that RB-CQL and CQL use identical hyper-parameters in all imbalanced datasets. The details of important hyper-parameters $\alpha$ of TD3+BC and CQL (RB-CQL) are shown in \Cref{tab:alpha}. and other hyper-parameters of TD3+BC and CQL (RB-CQL) are shown in \Cref{table:td3_hyp} and \Cref{table:cql_hyp}, where we note that the hyper-parameters in CQL (RB-CQL) with AntMaze tasks are suggested by the open-source code from \href{https://github.com/young-geng/CQL}{https://github.com/young-geng/CQL}. 

\begin{table}[h]
\centering
\caption{$\alpha$ used for TD3+BC, CQL, and RB-CQL.}
\label{tab:alpha}
\begin{tabular}{ll|rr} 
\toprule
\multicolumn{2}{c|}{Env}                     & \multicolumn{1}{c}{TD3+BC} & \multicolumn{1}{c}{CQL (RB-CQL)}  \\ 
\hline
\multirow{4}{*}{AntMaze-Meidum}  & Easy      & 2                          & 5                                 \\
                                 & Medium    & 2                          & 5                                 \\
                                 & Hard      & 2                          & 5                                 \\
                                 & Hard+     & 2                          & 5                                 \\
\multirow{4}{*}{AntMaze-Large}   & Easy      & 2                          & 5                                 \\
                                 & Medium    & 2                          & 5                                 \\
                                 & Hard      & 2                          & 5                                 \\
                                 & Hard+     & 2                          & 5                                 \\
\multirow{3}{*}{Mujoco-hopper}   & 95 random & 5                          & 20                                \\
                                 & 97 random & 5                          & 20                                \\
                                 & 99 random & 5                          & 20                                \\
\multirow{3}{*}{Mujoco-walker2d} & 95 random & 2                          & 30                                \\
                                 & 97 random & 2                          & 30                                \\
                                 & 99 random & 2                          & 30                                \\
\bottomrule
\end{tabular}
\end{table}

\begin{table}[htb]
\centering
\caption{The other hyperparameters of TD3+BC.}
\vspace{5pt}
\begin{tabular}{c|ll}
\toprule
& Hyperparameter & Value \\
\midrule
\multirow{6}{*}{Architecture}         & Critic hidden dim    & 256        \\
                                      & Critic hidden layers & 2          \\
                                      & Critic activation function & ReLU \\
                                      & Actor hidden dim     & 256        \\
                                      & Actor hidden layers  & 2          \\
                                      & Actor activation function & ReLU \\
                                      \midrule
\multirow{9}{*}{TD3+BC Hyperparameters} & Optimizer & Adam~\citep{adam} \\
                                      & Critic learning rate & 3e-4 \\
                                      & Actor learning rate  & 3e-4 \\
                                      & Mini-batch size      & 256 \\
                                      & Discount factor      & 0.99 \\
                                      & Target update rate   & 5e-3 \\
                                      & Policy noise         & 0.2 \\
                                      & Policy noise clipping & (-0.5, 0.5) \\
                                      & Policy update frequency & 2 \\
\bottomrule
\end{tabular}
\label{table:td3_hyp}
\end{table} %

\begin{table}[htb]
\caption{The other hyperparameters of CQL (RB-CQL).}
\centering
\vspace{5pt}
\begin{tabular}{c|lll}
\toprule
& Hyperparameter & AntMaze & Mujoco\\
\midrule
\multirow{6}{*}{Architecture}         & Critic hidden dim    & 256  & 256      \\
                                      & Critic hidden layers & 3  & 2        \\
                                      & Critic activation function & ReLU  & ReLU\\
                                      & Actor hidden dim     & 256  & 256      \\
                                      & Actor hidden layers  & 3   & 3       \\
                                      & Actor activation function & ReLU & 256\\
\midrule
\multirow{16}{*}{CQL Hyperparameters} & Optimizer & Adam~\citep{adam} & Adam    \\
                                      & Critic learning rate & 3e-4   &  3e-4  \\
                                      & Actor learning rate  & 1e-4   & 3e-4   \\
                                      & Mini-batch size      & 256    & 256    \\
                                      & Discount factor      & 0.99   & 0.99    \\
                                      & Target update rate   & 5e-3  & 5e-3 \\
                                      & policy\_log\_std\_multiplier & 0 & 1.0 \\
                                      & bc\_epochs & 40 & 0 \\
                                      & reward\_scale & 10 & 1\\
                                      & reward\_bias & -5 & 0 \\
                                      & Target entropy       & -1 $\cdot$ Action Dim & -1 $\cdot$ Action Dim \\
                                      & Entropy in Q target  & True  & True \\ 
                                      & Lagrange             & True  & False    \\
                                      & orthogonal\_init      & True  & False    \\
                                      & Num sampled actions (during eval) & 10 & 10 \\
                                      & Num sampled actions (logsumexp) & 10 & 10\\
                                      
\bottomrule
\end{tabular}
\label{table:cql_hyp}
\end{table}

\end{document}